\documentclass[review]{elsarticle}

\usepackage{bm}
\usepackage{bbm}
\usepackage{lineno,hyperref}
\usepackage{mathtools}
\usepackage{graphicx}
\usepackage{multirow}
 \usepackage{textcomp}
 \usepackage{tikz}
 \usetikzlibrary{arrows,positioning}
\usepackage{makecell} 
\setcellgapes{5pt}

\usepackage{amsmath,amsfonts,amssymb}

\usepackage{lineno,hyperref}
\modulolinenumbers[5]

\journal{}

\begin{document}
\fontsize{9}{8}\selectfont
\sloppy

\begin{frontmatter}

\title{Analyzing categorical time series with the R package \textit{ctsfeatures}}


\author{\'Angel L\'opez-Oriona\corref{mycorrespondingauthor}}
\ead{oriona38@hotmail.com, a.oriona@udc.es}

\author{Jos\'e A. Vilar}
\ead{jose.vilarf@udc.es}

\cortext[mycorrespondingauthor]{Corresponding author}

\address{Research Group MODES, Research Center for Information and Communication Technologies (CITIC), University of A Coru\~na, 15071 A Coru\~na, Spain.}




\begin{abstract}
Time series data are ubiquitous nowadays. Whereas most of the literature on the topic deals with real-valued time series, categorical time series have received much less attention. However, the development of data mining techniques for this kind of data has substantially increased in recent years. The R package \textbf{ctsfeatures} offers users a set of useful tools for analyzing categorical time series. In particular, several functions allowing the extraction of well-known statistical features and the construction of illustrative graphs describing underlying temporal patterns are provided in the package. The output of some functions can be employed to perform traditional machine learning tasks including clustering, classification and outlier detection. The package also includes two datasets of biological sequences introduced in the literature for clustering purposes, as well as three interesting synthetic databases. In this work, the main characteristics of the package are described and its use is illustrated through various examples. Practitioners from a wide variety of fields could benefit from the valuable tools provided by \textbf{ctsfeatures}.
\end{abstract}

\begin{keyword}
\textbf{ctsfeatures}, R package, categorical time series, feature extraction, association measures. 
\end{keyword}

\end{frontmatter}

\section{Introduction}\label{sectionintroduction}

Traditionally, most of the literature on time series analysis has focused on real-valued time series, while the study of time series with discrete response has received much less attention even though they arise in a natural way in many applications. Examples include weekly counts of new infections with a specific disease \cite{weiss2014binomial}, temporal records of EEG sleep states \cite{stoffer2000spectral}, the stochastic modeling of DNA sequence data \cite{fokianos2003regression, weiss2008measuring} and financial time series \cite{moysiadis2014binary}, and the use of hidden Markov processes to deal with protein sequences \cite{krogh1994hidden}, among others. An excellent introduction to the topic of discrete-valued time series is provided by \cite{weiss2018introduction}, including classical models, recent advances, relevant references and specific application areas. 

Categorical time series (CTS) are characterized by taking values on a qualitative range consisting of a finite number of categories, which is referred to as ordinal range, if the categories exhibit a natural ordering, or nominal range, otherwise. In this work, the most general case of nominal range is considered. Indeed, dealing with unordered qualitative outcomes implies that some classical analytic tools are not longer useful since categories cannot be ranked and no algebraic operations can be performed on them. This way, standard measures of location (mean, median, quantiles), dispersion (standard deviation, range) and serial dependence (autocorrelation, partial autocorrelation) are not defined in the nominal setting, and alternative quantities considering the qualitative nature of the range are needed for the analysis of CTS.

While early articles about CTS mainly focused on problems of statistical inference and modeling, several works addressing the construction of machine learning algorithms for categorical series have been published in recent years. For instance, in the context of clustering, \cite{cadez2003model} proposed an algorithm based on a mixture of first order Markov models to group time series recording navigation patterns 
of users of a web site. A different strategy was considered in \cite{garcia2015framework}, where a dissimilarity measure combining both closeness of raw categorical values and proximity between dynamic behaviors is used as input to a modified version of the $K$-modes algorithm. A feature-based clustering algorithm aimed at detecting groups of series with similar dependence structures was introduced by \cite{lopez2023hard}. Their approach consists of replacing the original series by a set of informative statistics extracted from each one of them, which are then used as input to a standard clustering procedure. The feature-based approach has also been employed for alternative tasks as CTS classification. For example, \cite{li2022interpretable} proposed a novel approach relying on the so-called spectral envelope and its corresponding set of optimal scalings, which characterize oscillatory patterns in a categorical series. Besides clustering and classification, there are other interesting and challenging problems related to categorical sequences, as anomaly detection \cite{horak2022nlp}. Previous references highlight the remarkable growth that CTS mining has recently undergone.   

According to these considerations, it seems clear that the development of software specifically designed to deal with CTS is of great importance. However, although there exist several packages revolving around the analysis of categorical data without a temporal nature (e.g., the R \cite{rsoftware} package \textbf{confreq} \cite{confreq, heine2021analysis} or the Python \cite{10.5555/1593511} package \textbf{categorical-encoding}), only a few computing tools are focused on categorical sequences. Moreover, these tools are often restricted to one specific task, application domain or class of categorical models. For instance, the R package \textbf{TraMineR} \cite{gabadinho2011analyzing} implements a collection of dissimilarity measures between CTS (see \cite{studer2016matters}) related to the so-called sequence analysis \cite{liao2022sequence}, a topic of great interest in social sciences. Most of these metrics attempt to detect differences concerning the order in which successive categories appear, the timing and the duration of the spells for different categories. Another interesting tool devoted to sequence analysis is the STATA \cite{statacorpstata} package SADI \cite{halpin2017sadi}, which incorporates graphical summaries and tools for cluster analysis besides computing several dissimilarity criteria. Some other packages are designed to deal with a particular type of models for CTS. For example, the R package \textbf{HMM} provides several functions to perform statistical inference for Hidden Markov Models (HMM). Note that, in terms of exploratory analyses, focusing on a specific class of models is too restrictive, since most real CTS datasets are expected to contain series generated from different types of stochastic processes. Although their usefulness is beyond doubt, none of the available software packages for handling CTS provide the necessary tools to compute classical statistical features for categorical series. 

The goal of this article is to present the R package \textbf{ctsfeatures}, which contains several functions to compute well-known statistical features for CTS. Besides providing a valuable description of the structure of the series, these features can be used as input for traditional machine learning procedures, as clustering, classification and outlier detection algorithms. In addition, \textbf{ctsfeatures} includes some visualization tools, as well as serial dependence plots which can be seen as an extension of the autocorrelation plots for numerical time series. The two databases of biological sequences introduced in \cite{lopez2023hard} are also available in the package along with several synthetic datasets formed by CTS generated from different stochastic processes. These data collections allow the users to test the functions provided in the package. 

In summary, \textbf{ctsfeatures} intends to integrate a set of simple but powerful functions for the statistical analysis of CTS into a single framework. Implementation was carried out by using the open-source R programming language due to the role of R as the most used programming language for statistical computing. The package \textbf{ctsfeatures} is  available from the Comprehensive R Archive Network (CRAN) at \url{https://cran.r-project.org/package=ctsfeatures}.  

The rest of the paper is structured as follows. A summary of relevant features to analyze marginal properties and serial dependence of categorical time series is presented in Section~\ref{sectionCTSfeatures}. Furthermore, features measuring cross-dependence between categorical and numerical processes are also introduced. The main functions implemented in \textbf{ctsfeatures} and the available datasets are listed and discussed in Section~\ref{sectionmainfunctions}. Section~\ref{sectionillustration} is devoted to illustrate the functionality of the package through several examples considering the synthetic data and the biological databases included in \textbf{ctsfeatures}. In addition, the way in which the outputs of some functions can be used to perform classical machine learning tasks is properly described. Lastly, some concluding remarks are provided in Section~\ref{sectionconcludingremarks}. 

\section{Analyzing marginal properties and serial dependence of categorical time series} \label{sectionCTSfeatures}
Let $\{X_t, t \in \mathbb{Z}\}$ be a categorical stochastic process taking values on a finite number $r$ of unordered qualitative categories, which are coded with the integer numbers from 1 to $r$. The range of the process is denoted by $\mathcal{V}=\{1,\ldots,r\}$. Hereafter, we assume that $X_t$ is bivariate stationary, that is the pairwise joint distribution of $(X_{t}, X_{t-l})$ is invariant in $t$ for arbitrary $l$ \cite{weiss2008measuring}. 

For $j=1,\ldots,r$, denote by $\bm e_j$ the unit vector of length $r$ taking the value 1 in its $j$th position and 0 in all others. Then, a useful equivalent representation of $\{X_t, t \in \mathbb{Z}\}$ is given by the multivariate process $\{\bm Y_t=(Y_{t,1}, \ldots, Y_{t,r})^\top, t \in \mathbb{Z}\}$ defined by $\bm Y_t=\bm e_j$ if $X_t=j$, which is usually referred to as the binarization of $\{X_t, t \in \mathbb{Z}\}$.

 The marginal distribution of $X_t$ is denoted by $\bm{p}=(p_1, \ldots, p_r)^\top$, with $p_j=P(X_t=j)=P(\bm Y_t=\bm e_j)$, $j=1,\ldots,r$. As the range of $\{X_t, t \in \mathbb{Z}\}$ is intrinsically qualitative, the expectation of the process can not be computed in the standard way. However, the marginal probabilities can be used to define several measures of categorical dispersion. Three of these measures are the Gini index, entropy and Chebycheff dispersion, which are given in the first rows of Table~\ref{tablefeaturescategoricalprocess}. The three measures have range $[0, 1]$, reaching the extreme values 0 and 1 in the case of a one-point distribution (minimum dispersion) and a uniform distribution (maximum dispersion), respectively. 

\begin{table}[ht]
	\centering 
	\resizebox{12cm}{!}{\begin{tabular}{llll} \hline 
 			\multicolumn{4}{c}{Univariate categorical processes}  \\  
			Measure & Definition & Range & Type \\ \hline
			Gini index &    $g=\frac{r}{r-1}(1-\sum_{i=1}^{r}p_i^2)$        &   $[0, 1]$    &  Marginal   \\  
			Entropy &     $e=\frac{-1}{\ln(r)}\sum_{i=1}^{r}p_i\ln p_i$       &   $[0, 1]$    &  Marginal   \\  
			Chebycheff dispersion &     $c=\frac{m}{m-1}(1-\max_ip_i)$       &   $[0, 1]$    &  Marginal   \\  
			Goodman and Kruskal\textquotesingle s $\tau$ 	& $\tau(l)=\frac{\sum_{i,j=1}^{r}\frac{p_{ij}(l)^2}{p_j}-\sum_{i=1}^rp_i^2}{1-\sum_{i=1}^rp_i^2}$           &    $[0, 1]$   &  Serial    \\
			Goodman and Kruskal\textquotesingle s $\lambda$	&      $\lambda(l)=\frac{\sum_{j=1}^{r}\max_ip_{ij}(l)-\max_ip_i}{1-\max_ip_i}$      &   $[0, 1]$    &   Serial   \\
			Uncertainty coefficient	&     $u(l)=-\frac{\sum_{i, j=1}^{r}p_{ij}(l)\ln\big(\frac{p_{ij}(l)}{p_ip_j}\big)}{\sum_{i=1}^{r}p_i\ln p_i}$       &   $[0, 1]$    &  Serial    \\
			Pearson measure	&    $\text{X}_n^2(l)=n\sum_{i,j=1}^{r}\frac{(p_{ij}(l)-p_ip_j)^2}{p_ip_j}$        &   $[0, n(r-1)]$    &   Serial   \\
			$\Phi^2$-measure	&       $\Phi^2(l)=\frac{\text{X}_n^2(l)}{n}$     &   $[0, r-1]$    &  Serial   \\
			Sakoda measure	&     $p^*(l)=\sqrt{\frac{r\Phi^2(l)}{(r-1)(1+\Phi^2(l))}}$       &   $[0, 1]$    &  Serial   \\
			Cramer\textquotesingle s $v$	&     $v(l)=\frac{\Phi(l)}{\sqrt{r-1}}$       &  $[0, 1]$     &  Serial   \\
			Cohen\textquotesingle s $\kappa$ 	&     $\kappa(l)=\frac{\sum_{j=1}^{r}(p_{jj}(l)-p_j^2)}{1-\sum_{i=1}^rp_i^2}$       &   $[-\frac{\sum_{i=1}^{r}p_i^2}{1-\sum_{i=1}^{r}p_i^2}, 1]$    &  Serial   \\
			Total correlation	&       $\Psi(l)=\frac{1}{r^2}\sum_{i,j=1}^{r}\psi_{ij}(l)^2$     &    $[0, 1]$   &  Serial   \\ \hline 
		\multicolumn{4}{c}{Bivariate processes with numerical and categorical components}  \\    
			Measure & Definition & Range & Type \\ \hline
			Total mixed cross-correlation	&    $\Psi^*_1(l)=\frac{1}{r}\sum_{i=1}^{r}\psi^*_{i}(l)^2$       &   $[0, 1]$    &  Serial   \\
			Total mixed q-cross-correlation &     $\Psi^*_2(l)=\frac{1}{r}\sum_{i=1}^{r}\int_{0}^{1}\psi^\rho_{i}(l)^2d\rho$       &   $[0, 1]$    &  Serial   \\ \hline 
	\end{tabular}}
	\caption{Some relevant features to measure dispersion and serial dependence of a categorical stochastic process (top) and cross-dependence between the categorical and numerical components of a bivariate mixed process (bottom).}
	\label{tablefeaturescategoricalprocess} 
\end{table}

To analyze the serial dependence structure of $\{X_t, t \in \mathbb{Z}\}$ for a given lag $l \in \mathbb{Z}$, it is interesting consider the lagged joint probabilities $p_{ij}(l)$ and the lagged conditional probabilities $p_{i|j}(l)$, for $i,j= 1, \ldots, r$, which are defined by
\begin{equation}\label{jointandconditionalp}
\begin{split}
p_{ij}(l)&=P(X_t=i, X_{t-l}=j)=P(\bm Y_t=\bm e_i, \bm Y_{t-l}=\bm e_j),  \\ p_{i|j}(l)&=P(X_t=i|X_{t-l}=j)=P(\bm Y_t=\bm e_i|\bm Y_{t-l}=\bm e_j)=\frac{p_{ij}(l)}{p_j}.
\end{split}
\end{equation}

Based on probabilities in \eqref{jointandconditionalp}, \cite{weiss2008measuring} introduce notions of perfect serial independence and dependence for a categorical process. Specifically, we have perfect serial independence at lag $l \in \mathbb{N}$ if and only if $p_{ij}(l)=p_ip_j$ for any $i,j \in \mathcal{V}$. On the other hand, we have perfect serial dependence at lag $l \in \mathbb{N}$ if and only if the conditional distribution $p_{\cdot|j}(l)$ is a one-point distribution for any $j \in \mathcal{V}$. This way, knowledge about $X_{t-l}$ does not help at all in predicting the value of $X_{t}$ in a perfect serially independent process, while $X_{t}$ is completely determined from $X_{t-l}$ under perfect serial dependence. Note that these concepts describe the so-called unsigned dependence of the process $X_{t}$. However, it is often desirable to know whether a process tends to stay in the state it has reached (positive dependence) or, on the contrary, the repetition of the same state after $l$ steps is infrequent (negative dependence). In other terms, by analogy with the autocorrelation function of a numerical process, which takes positive or negative values, it is interesting to introduce a signed dependence measure for categorical processes. Again following \cite{weiss2008measuring}, provided that perfect serial dependence holds, we have perfect positive (negative) serial dependence if $p_{i|i}(l)=1$ ($p_{i|i}(l)=0$) for all $i \in \mathcal{V}$.

An alternative way of evaluating serial dependence within a categorical process is by considering pairwise correlations in the binarization of  $X_{t}$ \cite{lopez2023hard}, that is, obtaining 
\begin{equation}\label{correlations}
\psi_{ij}(l)=Corr(Y_{t, i}, Y_{t-l, j})=\frac{p_{ij}(l)-p_ip_j}{\sqrt{p_i(1-p_i)p_j(1-p_j)}}.
\end{equation}

By construction, features in \ref{correlations} play a similar role as  the autocorrelation function of a numerical stochastic process, which accounts for their capability to characterize the underlying dependence. In fact, a clustering algorithm based on the values $\psi_{ij}(l)$ was proposed by \cite{lopez2023hard} and exhibited a high accuracy grouping together series with similar underlying dependence patterns. 

A range of association measures describing the serial dependence structure of a categorical process at lag $l$ can be defined by using the probabilities introduced in \eqref{jointandconditionalp} and \eqref{correlations}. In most cases, these measures evaluate the degree of deviation between a given quantity and its simplification under the assumption of serial independence. Hence, the value $0$ is associated with a serially independent process. Some of the most important association measures available in the literature are listed in the upper part of Table~\ref{tablefeaturescategoricalprocess}. For instance, the Pearson measure computes the squared differences between the joint probabilities $p_{ij}(l)$ and their corresponding factorization under independence, $p_ip_j$. Note that only Cohen\textquotesingle s $\kappa$ can take negative values since the differences between probabilities in the numerator are not squared. Indeed, $\kappa(l)$ takes its lowest (highest) possible value in the case of a perfect negative (positive) dependent process. 

In practice, the quantities in the upper part of Table~\ref{tablefeaturescategoricalprocess} must be estimated from a $T$-length realization of the process, $\overline{X}_t=\{\overline{X}_1,\ldots, \overline{X}_T\}$. For each measure, a standard estimate is directly obtained by replacing in the corresponding formulae the probabilities $p_i$ and $p_{ij}(l)$ by their respective unbiased estimates $\widehat{p}_i$ and $\widehat{p}_{ij}(l) $ given by 
\begin{equation} \label{estimates}
\begin{split}
\widehat p_i=\frac{N_i}{T}, & \, \text{ with } \, 
N_i = \sum_{t=1}^T \mathbbm{1}\left(\overline{X}_t=i\right),  \\
\widehat p_{ij}(l)= \frac{N_{ij}(l)}{T-l}, & \,
 \, \text{ with } \, N_{ij}(l)= 
\sum_{t=l+1}^T \mathbbm{1}\left( (\overline{X}_t,\overline{X}_{t-l})=(i,j) \right), 
\end{split}
\end{equation}
for $i,j=1,\ldots,r$, and where $\mathbbm{1}({\cal A})$ stands for the indicator function taking the value 1 if ${\cal A}$ happens and 0 otherwise. In all cases, the symbol $\widehat{\cdot}$ is used to denote the corresponding estimates (e.g., $\widehat{g}$ refers to the estimate of the Gini index). These estimates are often used to construct statistical procedures aimed at testing the serial independence of a given categorical process at a given lag $l$ (see Section~\ref{subsectionvisualizationtools}).
 
Another interesting issue when dealing with a categorical process $\{X_t, t \in \mathbb{Z}\}$ is to measure its degree of cross-dependence with a given real-valued process. Let $\{Z_t, t \in \mathbb{Z}\}$ be a strictly stationary real-valued process with variance $\sigma^2$. Then, for $i=1,\ldots,r$, the level of linear dependence between the $i$th category of $X_t$ and the process $Z_t$ at a given lag $l \in \mathbb{Z}$ can be evaluated by means of
\begin{equation}\label{totalmixedcorrelation}
\psi_{i}^*(l)=Corr(Y_{t,i}, Z_{t-l})=\frac{Cov(Y_{t,i}, Z_{t-l})}{\sqrt{p_i(1-p_i)\sigma^2}},
\end{equation}
where $Y_{t,i}$ is the $i$th component of the binarization of $X_t$. 

Cross-correlations $\psi_{i}^*(l)$ are based on expected values and constitute the traditional way to examine linear dependence. However, uncorrelated variables might exhibit dependence in different parts of the joint distribution. A deeper analysis of cross-dependence can be performed by examining dependence in quantiles. More specifically, by using the quantities
\begin{equation}\label{totalmixedqcorrelation}
	\psi_{i}^\rho(l)=Corr\big(Y_{t,i}, \mathbbm{1} \left( Z_{t-l}\leq q_{Z_t}(\rho)\right) \big)=\frac{Cov(Y_{t,i}, \mathbbm{1} \left(Z_{t-l}\leq q_{Z_t}(\rho)\right)}{\sqrt{p_i(1-p_i)\rho(1-\rho)}},
\end{equation}
for $i=1,\ldots, r$, where $\rho \in (0, 1)$ is an arbitrary probability level and $q_{Z_t}(\cdot)$ denotes the quantile function of the process $Z_t$. By considering different values for $\rho$, $\psi_{i}^\rho(l)$ provide measures of dependence at lag $l$ between the $i$th category of $X_t$ and the probabilities to exceed a range of quantiles of $Z_t$, which leads to a more comprehensive picture of the underlying dependence and allows us to detect different types of dependence structures.

In order to have available a global measure of linear cross-correlation between categorical and numerical processes, the features $\psi_{i}^*(l)$ can be used to define the \textit{total mixed cross-correlation at lag $l$} given by
\begin{equation}
\Psi^*_1(l)=\frac{1}{r}\sum_{i=1}^{r}\psi^*_{i}(l)^2.
\end{equation} 

Analogously, a measure of the quantile cross-correlation between both processes, herein referred to as \textit{total mixed q-cross-correlation at lag $l$}, can be defined as
\begin{equation}
	\Psi^*_2(l)=\frac{1}{r}\sum_{i=1}^{r}\int_{0}^{1}\psi^\rho_{i}(l)^2d\rho.
\end{equation} 

By definition, both measures $\Psi^*_1(l)$ and $\Psi^*_2(l)$ (see the lower part of Table~\ref{tablefeaturescategoricalprocess}) take values on $[0, 1]$, reaching the value 0 in the case of null cross-dependence between $X_t$ and $Z_t$, and larger values when a stronger degree of cross-dependence between both processes is present. 

In practice, given a $T$-length realization $(\overline{X}_t, \overline{Z}_t) =\{(\overline{X}_1, \overline{Z}_1),\ldots, (\overline{X}_T, \overline{Z}_T)\}$ of the bivariate process $\{(X_t, Z_t), t \in \mathbb{Z}\}$, respective estimates of $\psi^*_i(l)$ and $\psi^p_i(l)$ can be constructed by considering
\begin{equation}\label{estimateswithcov}
\begin{split}
\widehat{\psi}^*_i(l)&=\frac{\widehat{Cov}(Y_{t,i}, Z_{t-l})}{\sqrt{\widehat{p}_i(1-\widehat{p}_i)\widehat{\sigma}^2}}, \\
\widehat{\psi}^p_i(l)&=\frac{\widehat{Cov}(Y_{t,i}, I(Z_{t-l}\leq \widehat{q}_{Z_t}(\rho))}{\sqrt{\widehat{p}_i(1-\widehat{p}_i)\rho(1-\rho)}},
\end{split}
\end{equation} 
\noindent where $\widehat{Cov}(\cdot, \cdot)$ denotes the standard estimate of the covariance between two random variables, and $\widehat{\sigma}^2$ and $\widehat{q}_{Z_t}(\cdot)$ are standard estimates of the variance and the quantile function of process $Z_t$ computed from the realization $\overline{Z}_t$. Then, estimates in \eqref{estimateswithcov} allow to estimate the global measures $\Psi^*_1(l)$ and $\Psi^*_2(l)$ by computing 
$\widehat{\Psi}^*_1(l)=\frac{1}{r}\sum_{i=1}^{r}\widehat{\psi}^*_{i}(l)^2$ and $\widehat{\Psi}^*_2(l)=\frac{1}{r}\sum_{i=1}^{r}\int_{0}^{1}\widehat{\psi}^\rho_{i}(l)^2d\rho$, respectively.

So far, the features introduced to characterize dependence are defined in the time domain, but valuable features in the frequency domain have been also proposed in the literature. A well-known spectral feature is the so-called {\textit{spectral envelope}} \cite{stoffer1993spectral, stoffer2000spectral}. The idea is to assign numerical values ({\textit{scaling}}) to each of the categories of the process and then perform a spectral analysis of the resulting discrete-valued time series. Each scaling is properly selected to emphasize a specific cyclic pattern existing in the categorical process. Specifically, for $\boldsymbol \gamma=(\gamma_1, \ldots, \gamma_r)^\top \in \mathbb{R}^r$,  $X_t(\boldsymbol \gamma)=\boldsymbol \gamma^\top \boldsymbol Y_t$ denotes the real-valued stationary time series corresponding to the scaling that assigns the $i$th category of $X_t$ the value  $\gamma_i$, $i=1,\ldots,r$. For a given frequency $\omega$, the question is now to determine the ``most striking'' $\boldsymbol \gamma=\boldsymbol \gamma(\omega)$ (in some sense). For this purpose, \cite{stoffer1993spectral, stoffer2000spectral} apply a Fourier transform and compute the spectral density $f(\omega; \boldsymbol \gamma)$, or a sample version of it for given time series data. If $\sigma^2(\boldsymbol \gamma)$ denotes the variance of $\boldsymbol \gamma^\top \boldsymbol Y_t$, then $\boldsymbol \gamma (\omega)$ is chosen to maximize $f(\omega; \boldsymbol \gamma)/\sigma^2(\boldsymbol{\gamma})$. The corresponding maximal value, so-called $\lambda(\omega)$, is called the spectral envelope of process $X_t$. Note that $\lambda(\omega)$ expresses the maximal proportion of the variance that can be explained by the frequency $\omega$, and this maximal proportion is reached if the optimal scaling $\boldsymbol{\gamma}(\omega)$ is used. More details on the computation of $\lambda(\omega)$ and on corresponding sample versions $\widehat{\lambda}(\omega)$ can be found in \cite{stoffer1993spectral, stoffer2000spectral, shumway2000time}. If $\widehat{\lambda}(\omega)$ is plotted against $\omega$, a visual frequency analysis of the categorical time series is possible. 

\section{Main functions in ctsfeatures}\label{sectionmainfunctions}

This section is devoted to present the main content of package \textbf{ctsfeatures}. First, the datasets available in the package are briefly described, and then the main functions of the package are introduced, including both graphical and analytical tools. 

\subsection{Available datasets in ctsfeatures}\label{subsectiondatasets}

The package \textbf{ctsfeatures} includes some CTS datasets that can be used to evaluate different machine learning algorithms or simply for illustrative purposes. Specifically, the available data collections consist of two databases of biological sequences introduced by \cite{lopez2023hard} and three synthetic datasets used in \cite{lopez2023hard} for the evaluation of clustering algorithms. All of them are briefly described below.

\begin{itemize}
\item \textbf{Biological datasets}. The first biological dataset contains the genome of 32 viruses pertaining to 4 different families. Each genome can be seen as a sequence consisting of the four DNA bases, namely adenine, guanine, thymine  and cytosine. Therefore, this database contains 32 CTS with four categories. In addition, it is assumed the existence of four underlying classes corresponding to the family of each virus. The second database includes 40 protein sequences. As proteins are constituted of 20 different amino acids, each sequence can be considered as a CTS with 20 categories. In this case, there exist two different classes in the collection, since each protein is present either in human beings or in variants of COVID-19 virus. 

\item \textbf{Synthetic datasets}. Each one of the synthetic datasets is associated with a particular model of categorical stochastic process, namely Markov Chains (MC), Hidden Markov Models (HMM) and New Discrete ARMA (NDARMA) processes for the first, second and third database, respectively. In all cases, the corresponding database contains 80 series with 3 categories and length $T=600$, which are split into 4 groups of 20 series each. All series were generated from the same type of process defining the dataset, but the coefficients of the generating model differ between groups. For instance, four distinct transition matrices were used to generate the 80 series forming the first dataset. The specific coefficients were chosen according to Scenarios~1, 2 and 3 in \cite{lopez2023hard} (see Section 3.1). 
\end{itemize}

It is worth highlighting that the biological sequences are one of the most common types of CTS arising in practice. In fact, some machine learning algorithms for CTS are specifically designed to deal with this class of series \cite{fitzgerald2004clustering, kassim2017classification}. Therefore, it is useful for the user to have collections of this kind of data available through \textbf{ctsfeatures}. On the other hand, the distinct classes forming the synthetic databases can be distinguished by means of both the marginal distributions and the serial dependence patterns. This property makes these datasets particularly suitable to evaluate the effectiveness of the features in Table~\ref{tablefeaturescategoricalprocess} for several machine learning problems. In particular, the usefulness of these features to carry out clustering and classification tasks (among others) in these databases is illustrated in Section~\ref{subsectiondataminingtasks}. Table~\ref{summarydatasets} contains a summary of the 5 datasets included in \textbf{ctsfeatures}.    

\begin{table}[ht]
\centering 
		\resizebox{11cm}{!}{\begin{tabular}{llccrc} \hline 
		Dataset & Object & No. Series & $T$ &$|\mathcal{V}|$ & No. Classes \\ \hline 
		DNA sequences	&    \textit{GeneticSequences}       & 32  & Variable & 4 &  4     \\
		Protein sequences	&         \textit{ProteinSequences}       & 40 &  Variable & 20 &  2       \\
		Synthetic I	&         \textit{SyntheticData1}      & 80 & 600 & 3 &  4     \\
		Synthetic II	&        \textit{SyntheticData2}   & 80 & 600 &  3 &  4        \\
		Synthetic III		&   \textit{SyntheticData3}     & 80 & 600 & 3 &   4           \\ \hline 
	\end{tabular}}
	\caption{Summary of the 5 datasets included in \textbf{ctsfeatures}.}
	\label{summarydatasets}
\end{table}


\subsection{Graphical tools in ctsfeatures}\label{subsectionvisualizationtools}

The set of visualization tools for CTS available in \textbf{ctsfeatures} is presented and properly detailed in this section. Specifically, graphs for exploratory purposes, two types of plots aimed at characterizing serial dependence, and two graphs mainly designed for quality control. 

\subsubsection{Time series plot}\label{subsubsectiontsp}

The most common type of graph when dealing with real-valued time series is probably the so-called time series plot, which displays the values of the time series ($y$-axis) with respect to time ($x$-axis). A similar representation can be indeed considered for categorical series. As the categories have been coded with integer numbers from 1 to $r$, the range is $\mathcal{V}=\{1, 2, \ldots, r\}$ and the classical time series plot can be easily depicted for any CTS. However, although this plot can be useful for exploratory purposes, it presents serious problems with nominal time series because a comparison of the different categories in a numerical scale is not possible in this case. Note that this issue not longer exists by analyzing ordinal time series. In fact, as the range of these series exhibits a natural ordering, the corresponding plot is meaningful as long as this ordering is preserved when encoding the categories, i.e., $1 \le 2 \le...\le r$. 

\begin{figure}[ht]
	\centering
	\includegraphics[width=0.9\textwidth]{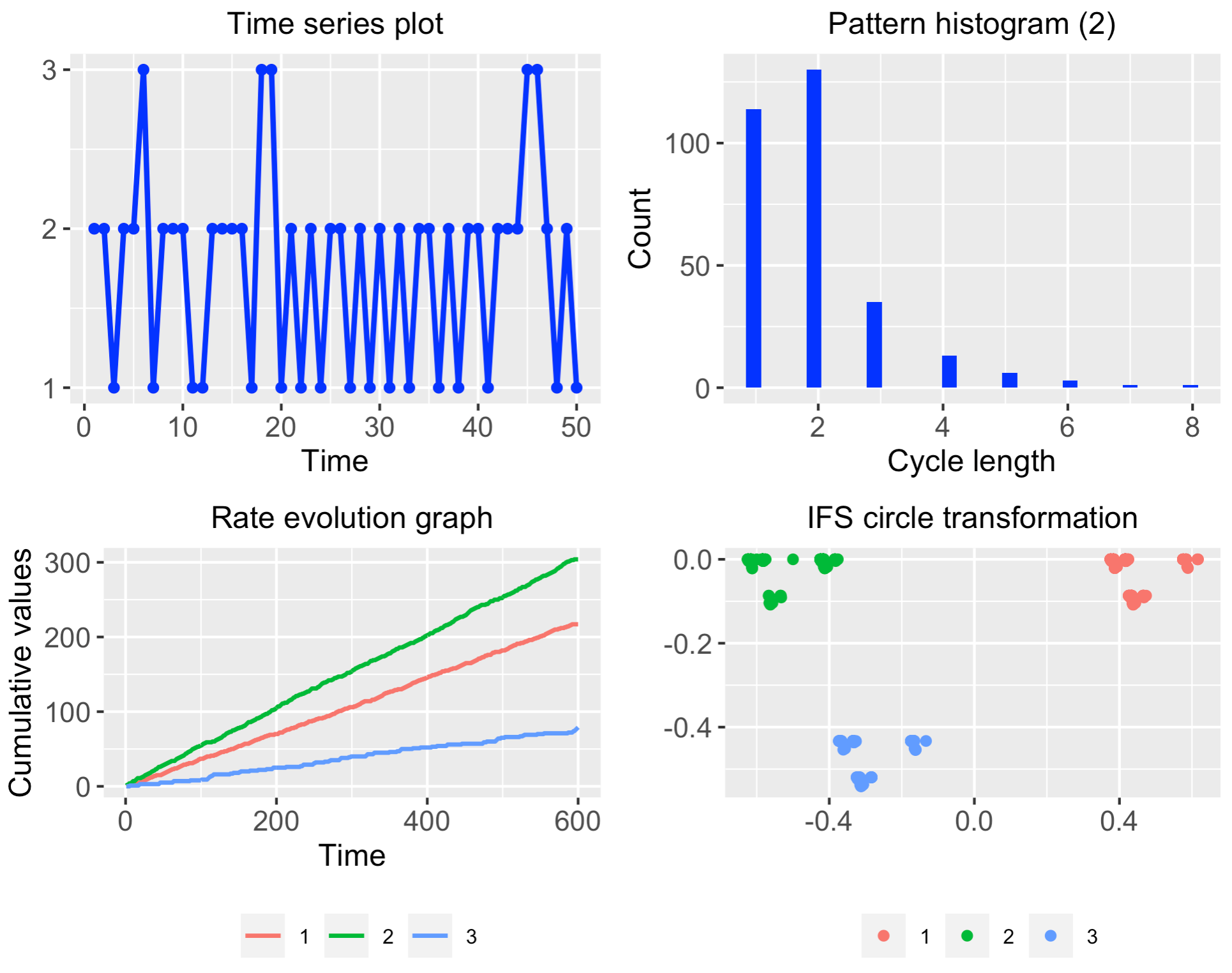}
	\caption{Several visualizations for the first series in dataset \textit{SyntheticData1}.}
	\label{v1}
\end{figure}

The top left panel of Figure~\ref{v1} shows the time series plot based on the first 50 observations of the first series in \textit{SyntheticData1}. As this series was generated from a MC, it has a nominal nature, so the corresponding categories were arbitrarily coded by using the integers 1, 2 and 3. 

\subsubsection{Rate evolution graph}\label{subsubsectionreg}

Since the standard time series plot cannot be applied to nominal time series in a completely meaningful way, several alternative tools for a visual analysis of CTS have been introduced in the literature (see \cite{weiss2008visual} for a survey). Although none of these graphs seems to be a perfect substitute for the time series plot in the categorical setting, the so-called rate evolution graph proposed by \cite{ribler1997visualizing} is one of the most frequently used tools for stationary analysis \cite{weiss2018introduction}. Let $\overline{\boldsymbol Y}_t=\{\overline{\boldsymbol Y}_1, \ldots, \overline{\boldsymbol Y}_T\}$ with $\overline{\boldsymbol Y}_k=(\overline{Y}_{k,1}, \ldots, \overline{Y}_{k,r})^\top$ such that $\overline{Y}_{k,i}=1$ if $\overline{X}_k=i$ ($k=1,\ldots,T$, $i=1,\ldots,r$) be the binarized time series associated with the $T$-length categorical series $\overline{X}_t$. Consider the series of cumulated sums given by $\overline{\boldsymbol C}_t=\{\overline{\boldsymbol C}_1, \ldots, \overline{\boldsymbol C}_T\}$ with $\overline{\boldsymbol C}_k=\sum_{s=1}^{k}\overline{\boldsymbol Y}_s$, $k=1,\ldots,T$. The rate evolution graph displays a standard time series plot for each one of the components of $\overline{\boldsymbol C}_t$ simultaneously in one graph. For each one of the components, the slope of the corresponding curve is an estimate for the associated marginal probability. If the process is stationary, then the curves should be approximately linear with $t$, while visible violations of linearity indicate nonstationarity. In fact, the rate evolution graph is frequently employed to check the assumption of stationarity before carrying out specific machine learning tasks in CTS databases (see, e.g., Section 6.1 in \cite{lopez2023hard}). 

The rate evolution graph corresponding to the first time series in dataset \textit{SyntheticData1} is shown in the bottom left panel of Figure \ref{v1}.

\subsubsection{Pattern histograms}\label{subsubsectionph}

A categorical series $\overline{X}_t$ can be described by means of the occurrence of certain patterns called cycles. Specifically, a cycle, beginning at time $t_1$ with $\overline{X}_{t_1}=j$, is closed at time $t_1+t_2>t_1$ if and only if the series returns to the value $j$ for the first time after $t_1$, i.e., if $\overline{X}_{t_1}=j=\overline{X}_{t_1+t_2}$ but $\overline{X}_{t^*} \ne j$ for all $t_1 < t^*<t_1+t_2$. By definition, a cycle starts at any time, and a previous cycle is closed except the case when the corresponding symbol occurs for the first time. The length of a cycle starting at $t_1$ and ending at $t_1+t_2$ is defined as $t_2$. Note that the cycles can be easily identified by examining the binarized time series. In fact, for a given category $j \in \mathcal{V}$, a cycle of length $t_2$ starting at time $t_1$ occurs if and only if $\overline{Y}_{t_1,j}=1=\overline{Y}_{t_1+t_2,j}$ but $\overline{Y}_{t^*} =0$ for all $t_1 < t^*<t_1+t_2$. The distribution of the length of the cycles existing in a given CTS represents an important statistical feature of the series. Thus, for each category $j \in \mathcal{V}$, \cite{weiss2008visual} suggests to record the corresponding cycles, group them according to their length, and construct an histogram with the corresponding counts. In this way, each series gives rise to $r$ histograms, which are often referred to as pattern histograms. Pattern histograms can be used for several purposes, including model identification, hypothesis testing or anomaly detection, among others. 

The top right panel of Figure~\ref{v1} contains the pattern histogram concerning the second category of the first series in dataset \textit{SyntheticData1}.  

\subsubsection{Circle transformation}\label{subsubsectionct}

Other type of exploratory tool for visualizing categorical time series is the so-called IFS circle transformation. This approach was suggested by \cite{weiss2005discover} and is based on iterated function systems (IFS). Although IFS are mainly known for their use to generate fractals, they are frequently applied for the visualization of genetic sequences, a particular type of CTS. The main idea behind the approach proposed by \cite{weiss2005discover} is that an arbitrary categorical series can be appropriately transformed by linear contractions. Afterwards, similar strings are represented by close points in $\mathbb{R}^2$. So pattern analysis can be done visually: the existence and frequency of patterns can be studied by zooming into certain regions of the real plane $\mathbb{R}^2$. The procedure by \cite{weiss2005discover} consists of the following steps: 
\begin{enumerate}
\item Transform the series range $\mathcal{V}$ by applying the circle transformation, which is a one-to-one mapping $\boldsymbol \varphi$ from $\mathcal{V}$ on points of the unit circle in $\mathbb{R}^2$ defined by
\begin{equation}
\boldsymbol \varphi(i)=\Bigg(\cos \bigg(\frac{2\pi(i-1)}{r}\bigg), \sin \bigg(\frac{2\pi(i-1)}{r}\bigg) \Bigg). 
\end{equation}
\item Construct the series $\overline{\boldsymbol R}_t=\{\overline{\boldsymbol R}_1, \ldots, \overline{\boldsymbol R}_T\}$ such that $\overline{\boldsymbol R}_k=\boldsymbol \varphi(\overline{X}_k)$, $k=1,\ldots,T$. 
\item Generate the fractal series, denoted by $\overline{\boldsymbol F}_t=\{\overline{\boldsymbol F}_1, \ldots, \overline{\boldsymbol F}_T\}$, by means of the following recursion:
\begin{equation}\label{fractalseries}
\overline{\boldsymbol F}_k=\alpha \overline{\boldsymbol F}_{k-1}+\beta \overline{\boldsymbol R}_k, 
\end{equation}
where $\alpha \in (0,1)$, $\beta>0$ and $\overline{\boldsymbol F}_0$ is an arbitrary initialization from $\mathbb{R}^2$. 

\end{enumerate}

A scatterplot of the two-dimensional fractal time series $\overline{\boldsymbol F}_t$ (IFS circle transformation) is frequently useful to identify cycles of arbitrary length (see Theorem 4.2 in \cite{weiss2008visual}). In fact, this graph contains frequency information on any subsequence of arbitrary length within it. The corresponding information can be obtained simply by zooming in into interesting regions. Interesting explanations concerning the interpretation of this visualization tool are given in Example 4.3 of \cite{weiss2008visual}. 

The IFS circle transformation for the first series in dataset \textit{SyntheticData1} is depicted in the bottom right panel of Figure \ref{v1}.  

\subsubsection{Serial dependence plots}\label{subsubsectionsdp}


When dealing with real-valued time series, a plot of the autocorrelation function (ACF) is usually depicted to identify the lags showing significant correlations and analyse the serial dependence structure. Since this function is not defined in the categorical case (neither nominal nor ordinal), alternative statistical representations are required. For a given CTS, a well-known approach to examine serial dependence at lag $l \in \mathbb{Z}$ consists in analyzing the sample versions of the Cramer\textquotesingle s $v$, $\hat{v}(l)$, and the Cohen\textquotesingle s $\kappa$, $\hat{\kappa}(l)$ (see Table~\ref{tablefeaturescategoricalprocess}). While the former statistic assesses the unsigned dependence, the later measures signed dependence (both dependence notions introduced in Section~\ref{sectionCTSfeatures}). In both cases, the value 0 is associated with an i.i.d. process. Moreover, the asymptotic distributions of the estimates $\widehat{v}(l)$ and $\widehat{\kappa}(l)$ under the i.i.d. assumption were derived by \cite{weiss2013serial} and \cite{weiss2011empirical}, respectively. The corresponding limit distributions are given by 
\begin{equation}\label{test1}
T(r-1)\widehat{v}(l)^2\underset{a}{\sim}\chi^2_{(r-1)^2} \quad \text{and} \quad \sqrt{\frac{T}{\widehat{ V}(\boldsymbol p)}}\bigg(\widehat{\kappa}(l)+\frac{1}{T}\bigg)\underset{a}{\sim}N\big(0, 1\big),
\end{equation}
where $\widehat{V}(\boldsymbol p)=1-\frac{1+2\sum_{i=1}^{r}\widehat{p}_i^3-3\sum_{i=1}^{r}\widehat{p}_i^2}{(1-\sum_{i=1}^{r}\widehat{p}_i^2)^2}$. 

The asymptotic results in \eqref{test1} are quite useful in practice, since they can be used to test the null hypothesis of serial independence at lag $l$. In particular, critical values for a given significance level $\alpha$ can be computed, which do not depend on the specific lag. Thus, in both cases, serial dependence plots can be constructed by following a similar approach as the ACF plot in the real-valued case. Specifically, after setting a maximum lag $L$ of interest, the values of $\widehat{v}(l)$ or $\widehat{\kappa}(l)$, for lags $l$ ranging from 1 to $L$, are simultaneously depicted in one graph. Next, the corresponding critical value is added to the plot by means of a horizontal line. According to \eqref{test1}, the critical values for an arbitrary significance level $\alpha$ are given by
\begin{equation}\label{test2}
+\sqrt{\frac{\chi^2_{(r-1)^2, 1-\alpha}}{T(r-1)}} \quad \text{and} \quad \pm \frac{z_{1-\alpha/2}}{\sqrt{T/V(\widehat{\boldsymbol p})}}-\frac{1}{T},
\end{equation} 
for the plots based on $\widehat{v}(l)$ and $\widehat{\kappa}(l)$, respectively, 
where $\chi^2_{g, \tau}$ and $z_{\tau}$ denote the respective $\tau$-quantiles of the $\chi^2$ distribution with $g$ degrees of freedom and the standard normal distribution. The described graphs allow easily to identify the collection of significant lags for a given categorical series. Likewise the autocorrelation plot in the numerical setting, serial dependence plots for CTS can be used for several purposes, including model selection or identification of regular patterns in the series, among others. 

\begin{figure}[ht]
	\centering
	\includegraphics[width=0.9\textwidth]{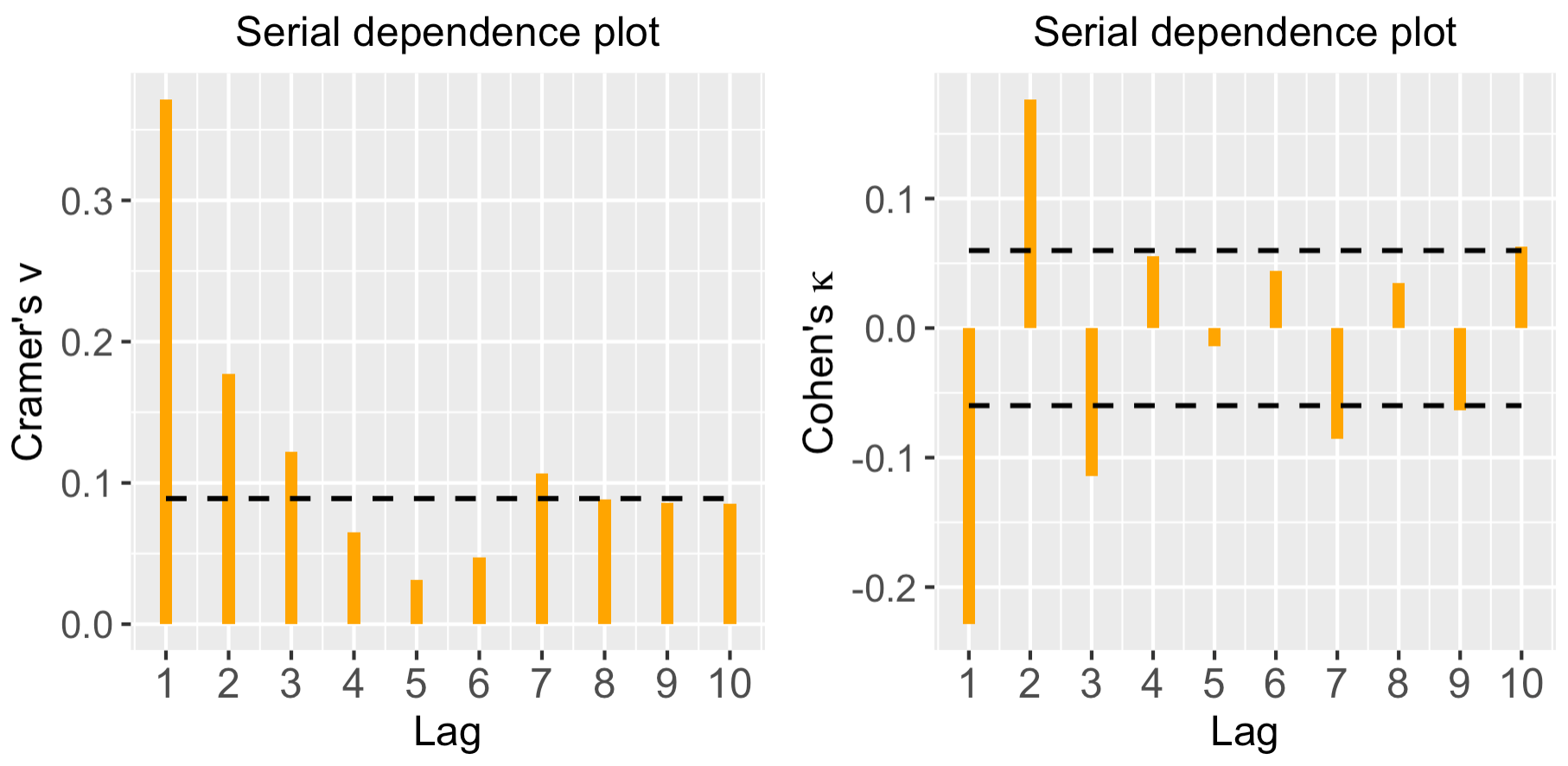}
	\caption{Serial dependence plots for the first series in dataset \textit{SyntheticData1}.}
	\label{v2}
\end{figure}

For illustrative purposes, the serial dependence plots based on $\widehat{v}(l)$ and $\widehat{\kappa}(l)$ for the first series in dataset \textit{SyntheticData1} are displayed Figure~\ref{v2}. A maximum lag $L=10$ was considered in both cases. 

\subsubsection{Control charts}\label{subsubsectioncc}

Techniques of statistical process control (SPC) aim at monitoring and improving production processes. Most popular among them are \textit{control charts}, which help to identify deviations from the state of control, but are also frequently used for exploratory purposes. In the latter context, the calculation of the so-called \textit{control limits} is based on historical data and a hypothetical process model. If some records violate these limits, they are flagged to be \textit{out of control} and possible causes for those deviations are analyzed. In particular, some of these records can be excluded if they are considered as single outliers. Afterwards, revised control limits are determined. However, if the out-of-control points imply a systematic deviation from the model assumptions, then the original model has to be rejected. 

As it is stated in \cite{weiss2008visual}, it is characteristic for categorical processes that the control statistics $C_t$ does not only depend on the monitoring time $t$, but also on the concrete value observed at time $t$, in the sense that the distribution of $C_t$ is influenced by the concrete outcome. Consider, for instance, the case that $C_t$ expresses the length of a cycle ending with $X_t=i$. Then, the corresponding cycle length distribution depends on $i$, leading to individual control limits $LCL_t$ and $UCL_t$ for $C_t$. If $\mu_t$ denotes the corresponding mean of $C_t$, we consider the standardized statistics $T_t=T_t^{(L)}+T_t^{(U)}$, where
\begin{equation}\label{equationcontrol1}
T_t^{(L)}=\min \left(0, \frac{C_t-\mu_t}{\left|L C L_t-\mu_t\right|}\right) \quad\text{and} \quad  T_t^{(U)}=\max \left(0, \frac{C_t-\mu_t}{\left|U C L_t-\mu_t\right|}\right).
\end{equation}

Thus, an out-of-control alarm is signaled if and only if $T_t<-1$ or $T_t>1$. 

An alternative possibility for controlling categorical processes relies on monitoring their marginal distribution. Specifically, if $\boldsymbol c \in [0,1]^r$ and $0<\lambda<1$, then 
\begin{equation}\label{equationcontrol2}
\widehat{\boldsymbol{\pi}}_t^{(\lambda)}=\lambda  \widehat{\boldsymbol{\pi}}_{t-1}^{(\lambda)}+(1-\lambda)  \boldsymbol{Y}_t, \, \, \text{for} \,\, t \geq 1, \quad \text{and} \quad \widehat{\boldsymbol{\pi}}_0^{(\lambda)}=\boldsymbol{c},
\end{equation} 
is called an EWMA estimator of the marginal distribution. If $\boldsymbol{c}$ is set equal to the hypothetical marginal distribution vector, then $\widehat{\boldsymbol \pi}_t^{(\lambda)}$ is unbiased in the state of control, and its variance $\sigma_{t, i}^2$ is an expression depending on the concrete process model. The exact distribution is difficult to derive, but if $\lambda$ is chosen large enough, then  $k \sigma$-limits will also perform well. So we choose $L C L_i=p_i-k \sigma_{t, i}$ and $U C L_i=p_i+k  \sigma_{t, i}$, and statistic \eqref{equationcontrol1} simplifies to 
\begin{equation}\label{equationcontrol3}
T_{t, i}=\frac{\hat{\pi}_{t, i}^{(\lambda)}-p_i}{k \cdot \sigma_{t,i}},
\end{equation}
where $\hat{\pi}_{t, i}^{(\lambda)}$ is the $i$th component of vector $\widehat{\boldsymbol{\pi}}_t^{(\lambda)}$. 

Finally, there are some modifications of the EWMA approach for controlling the marginal probabilities. For instance, if the range is very large, then it may be preferable not to plot all statistics $T_{t, i}$ given by the formula \eqref{equationcontrol3} simultaneously, but only the statistics
\begin{equation}\label{equationcontrol4}
T_t^{\min }=\min _{i \in \mathcal{V}} T_{t, i} \quad \text { and } \quad T_t^{\max }=\max _{i \in \mathcal{V}} T_{t, i}.
\end{equation}

\begin{figure}
	\centering
	\includegraphics[width=0.9\textwidth]{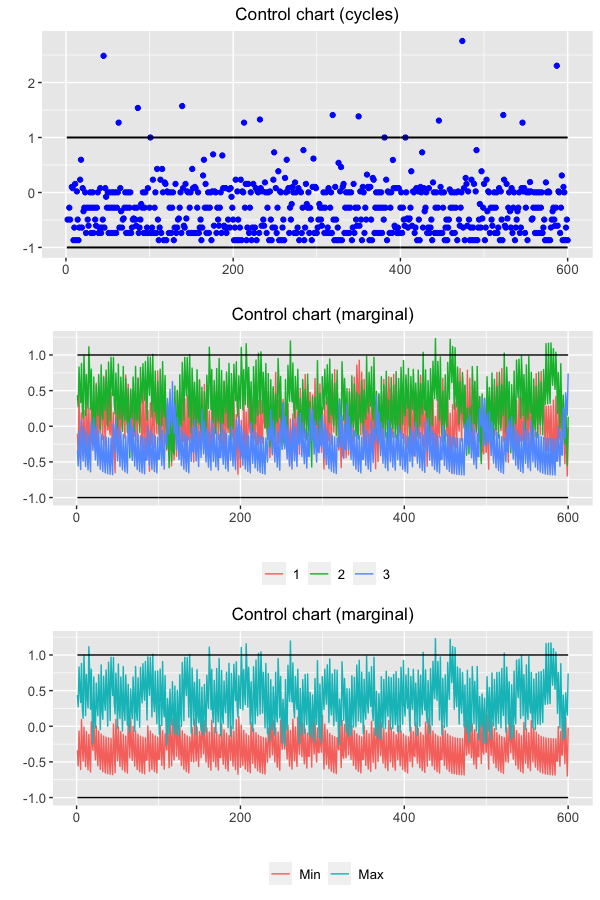}
	\caption{Control charts associated with expressions \eqref{equationcontrol1} (top panel), \eqref{equationcontrol2} (middle panel) and \eqref{equationcontrol4} (bottom panel) concerning the first time series in dataset \textit{SyntheticData1}.}
	\label{ccplots}
\end{figure}

The panels of Figure~\ref{ccplots} show control charts associated with expressions \eqref{equationcontrol1} (top), \eqref{equationcontrol2} (middle) and \eqref{equationcontrol4} (bottom) concerning the first time series in dataset \textit{SyntheticData1}. In all cases, some observations are beyond the upper control limit, thus deserving a careful investigation. 

A summary of the main functions for visualization available in \textbf{ctsfeatures} is provided in Table~\ref{tablefunctions1}.  

\begin{table}[ht]
\centering 
	\begin{tabular}{ll} \hline 
		Graph & Function in \textbf{ctsfeatures} \\ \hline 
		Time series plot	&    \textit{cts\_plot()}             \\
		Rate evolution graph	&         \textit{rate\_evolution\_graph()}             \\
		IFS circle transformation	&         \textit{ifs\_circle\_transformation()}             \\
		Pattern histogram	&         \textit{pattern\_histogram()}             \\
		Serial dependence plot \big($\widehat{v}(l)$\big) &      \textit{plot\_cramers\_vi()}   \\ 
		Serial dependence plot  \big($\widehat{k}(l)$\big) &      \textit{plot\_cohens\_kappa()}   \\ 
		Control chart (cycle length)		&         \textit{cl\_control\_chart()}             \\
		Control chart (marginal distribution)		&   \textit{marginal\_control\_chart()}           \\ \hline 
	\end{tabular}
	\caption{Functions for visualization in \textbf{ctsfeatures}.}
	\label{tablefunctions1}
\end{table}

\subsection{Functions for feature extraction in ctsfeatures} \label{subsectionfeaturesfunctions}

Package \textbf{ctsfeatures} includes a range of functions to compute well-known statistical quantities providing information on marginal and serial properties of CTS. All the functions are based on the estimated features presented in Section~\ref{sectionCTSfeatures} and are summarised in Table~\ref{tablefunctions2}. It is worth highlighting that most of them are not limited to obtain the estimated versions of the features in Table~\ref{tablefeaturescategoricalprocess}, but in addition the individual components of the corresponding estimates are also provided. For instance, the function \textit{cohens\_kappa()} computes by default the value of $\widehat{\kappa}(l)$ for a given lag $l$ which must be properly given as input. However, when the argument \textit{features=TRUE} is used, the function returns a vector formed by the estimates $\frac{p_{jj}(l)-p_j^2}{1-\sum_{i=1}^{r}p_i^2}$, $j=1,\ldots,r$. Note that, when the goal is to carry out a machine learning task as clustering or classification, usually the latter output is more useful than the former, since it provides a more meaningful description of the dependence structure of the corresponding CTS. The argument \textit{features=TRUE} has an analogous purpose for the majority of functions in Table~\ref{tablefunctions2}. Lastly it is worthy noting that computation of the estimated spectral envelope is performed by using the package \textbf{astsa} \cite{astsa}. 

\begin{table}[ht]
\centering 
	\begin{tabular}{llp{0.6cm}ll} \hline 
		Feature & Function \textbf{ctsfeatures} & &
        Feature & Function \textbf{ctsfeatures} 
        \\ \cline{1-2} \cline{4-5}
        $\widehat{g}$	 &  \textit{gini\_index()}   & & 
        $\widehat{p}^*(l)$	&  \textit{sakoda\_measure()} 
        \\
        $\widehat{e}$	&  \textit{entropy()} & &
        $\widehat{v}(l)$ &  \textit{cramers\_vi()} 
        \\
        $\widehat{c}$	& \textit{chebycheff\_dispersion()}  & &
        $\widehat{\kappa}(l)$	&   \textit{cohens\_kappa()}  
        \\
        $\widehat{\tau}(l)$	& \textit{gk\_tau()}  & &
        $\widehat{\Psi}(l)$	&  \textit{total\_cor()}  
       \\  
       $\widehat{\lambda}(l)$ & \textit{gk\_lambda()} & &
       $\widehat{\lambda}(\omega)$	& \textit{spectral\_envelope()}
        \\
       $\widehat{u}(l)$ &  \textit{uncertainty\_coefficient()} & &
       $\widehat{\Psi}^*_1(l)$	&  \textit{total\_mixed\_cor()}
       \\
       $\widehat{\text{X}}_n^2(l)$	& \textit{pearson\_measure()} & &
       $\widehat{\Psi}^*_2(l)$	&      \textit{total\_mixed\_qcor()} 
       \\ 
       $\widehat{\Phi}^2(l)$	&   \textit{phi2\_measure()} & &  
        &  
        \\       \hline 
	\end{tabular}
	\caption{Functions for feature extraction implemented in \textbf{ctsfeatures}.}
	\label{tablefunctions2}
\end{table}


\section{Using ctsfeatures: Generalities and illustrative examples}\label{sectionillustration}


\subsection{Some generalities about ctsfeatures}

The majority of functions in \textbf{ctsfeatures} take as input a single $T$-length CTS, $\overline{X}_T$, represented through a vector of equal length of class \textit{factor}, the standard R class for categorical objects. Factors have associated a vector of character strings, called \textit{levels}, where the categories forming the range of the CTS are stored and arranged in a given order, usually specified by the user.

The output of the graphic functions listed in Table~\ref{tablefunctions1} is simply the requested plot, which can be customised to some extent. Functions in Table~\ref{tablefunctions2} return by default the corresponding estimate for a given lag $l \in \mathbb{Z}$. However, as detailed in Section~\ref{subsectionfeaturesfunctions}, most of these functions admit the argument \textit{features=TRUE}, which produces a different outcome by returning a vector with the individual terms used to construct the estimate. It is worth remarking that most functions in \textbf{ctsfeatures} require the categories to be specified in the desired order. This is done by means of the argument \textit{categories}. In this way, several issues can be avoided. For instance, a particular realization does not necessarily include all the possible categorical values. Therefore, when analyzing such a series, one could ignore the existence of some categories by properly using the argument  \textit{categories}.

Every database included in \textbf{ctsfeatures} is stored as a list with the name indicated in the first column of Table~\ref{summarydatasets} and contains two elements, which are described below.
\begin{itemize}
	\item The element called \textit{data} is a list of matrices with the categorical series of the corresponding collection. 
	\item The element named \textit{classes} includes a vector of class labels associated with the objects in \textit{data}.
\end{itemize} 

Let\textquotesingle s visualize the first twenty observations of one series in the dataset \textit{GeneticSequences}.
\begin{verbatim}
> library(ctsfeatures)
> GeneticSequences$data[[1]][1:20]
[1] a t g g c c c a a g c a c a a a t t c t
Levels: a c g t
\end{verbatim}

In this series, the DNA bases, adenine, guanine, thymine  and cytosine, were coded with the letters `a', `g', `t' and `c', respectively. In addition, the categories were alphabetically ordered. Indeed, a different order could be specified, but one should be always aware of the considered order when analyzing the output of several functions (see Sections~\ref{subsectionvisualizing} and \ref{subsectiondataminingtasks}). 

\subsection{Visualizing categorical time series and testing for serial dependence}\label{subsectionvisualizing}

The functions described in Section~\ref{subsectionvisualizationtools} implementing graphical tools allow the user to construct valuable plots on the performance and dependence structure of a categorical time series. The classical plot showing the profile of the series is obtained with the function \textit{cts\_plot()}. For instance, a plot based on the first 50 observations of the first series in dataset \textit{SyntheticData1} is generated as follows.
\begin{verbatim}
> categories <- factor(c(1, 2, 3))
> cts_plot(series = SyntheticData1$data[[1]][1:50],
+          categories = categories)
\end{verbatim}

Note that the time series in dataset \textit{SyntheticData1} are nominal, taking values on three attributes coded with integers from 1 to 3. The desired order for the categorical range is indicated through the argument \textit{categories}. The corresponding output is depicted in the top left panel of Figure~\ref{v1} and valuable information can be extracted from this graph. For instance, it seems that the third category has a much lower marginal probability than the remaining. In addition, one could try to derive some conclusions about the serial dependence structure of the series. However, as stated in Section~\ref{subsubsectiontsp}, the time series plot can be misleading when visualizing nominal time series. In fact, the consideration of a numerical scale in the $y$-axis implicitly assumes an underlying ordering, which does not exist in the nominal setting. One of the most well-known proposal for the visual analysis of nominal time series is the rate evolution graph (see Section~\ref{subsubsectionreg}), which is obtained for the first series in \textit{SyntheticData1} by executing 
  \begin{verbatim}
> rate_evolution_graph(series = SyntheticData1$data[[1]], 
+                      categories = categories)
 \end{verbatim}
 
The above code produces the plot showed in the bottom left panel of Figure~\ref{v1}. This representation indicates a stationary behavior of the examined time series (at least concerning the marginal distribution), since the three curves exhibit an approximately linear behavior. The slopes of these curves give estimates of the corresponding marginal probabilities. Note that the slope of the blue line is substantially lower than the slopes of their counterparts, thus indicating that the third category appears with much lower frequency than the remaining ones, which is consistent with the time series plot in the top left panel of Figure~\ref{v1}.

While the rate evolution graph gives information about the marginal distribution, some alternative graphical representations are useful to analyze the existence of certain serial patterns, such as cycles. One of them is the pattern histogram (see Section~\ref{subsubsectionph}), which represents counts of the cycles associated with a given category $i \in \mathcal{V}$ after grouping them in accord with their length. Pattern histograms can be directly generated by using the function \textit{pattern\_histogram()}, whose main arguments are the corresponding CTS and the category for which we wish to analyze the cycles. In the following code, we indicate \textit{category=2}.
 \begin{verbatim}
> pattern_histogram(series = SyntheticData1$data[[1]], 
+                   category = 2)
 \end{verbatim}

The generated pattern histogram is shown in the top right panel of Figure~\ref{v1}. It is observed that the second category frequently repeats itself either immediately or after one step, as the cycles of lengths 1 and 2 are the most common. Note that large counts for these cycles are totally related to the fact that the second category is the one displaying the maximum marginal probability as seen in the rate evolution graph. Similar analyses could be carried out by considering the pattern histograms of the remaining two categories. 

Another interesting graph to examine the cyclical behavior of a categorical series is the circle transformation (see Section~\ref{subsubsectionct}). Frequency information concerning any subsequence in the original CTS can be extracted from this tool. Let\textquotesingle s construct the circle transformation of the series we are dealing with throughout this section. To this aim, we use the function \textit{ifs\_circle\_transformation()}. The parameters $\alpha$ and $\beta$ involved in the construction of the fractal time series in \eqref{fractalseries} must be given as input to this function. 
\begin{verbatim}
> ct <- ifs_circle_transformation(series = SyntheticData1$data[[1]], 
+                categories = categories, alpha = 0.17, beta = 0.10)
> ct
\end{verbatim}

We set the values $\alpha=0.17$ and $\beta=0.10$. The corresponding plot is provided in the bottom right panel of Figure~\ref{v1}. This graph is of limited use to perform a static analysis, but it is very helpful as an interactive exploratory tool, since the circles of any level are always organized in the same way. For instance, a subsequence starting with `1' is always at the top of its parent circle. In fact, by navigating appropriately within the graph, one could determine the frequency of any arbitrary string in the original series. Thus, to obtain the number of subsequences of the form `111', we should first zoom into the upper right part within the circle of red points. This results in another circle whose upper right part (which can be accessed again via zooming) contains as many points as subsequences of the desired type. The command \textit{coord\_cartesian()} of the R package \textbf{ggplot2} \cite{ggplot2} provides an easy way to navigate within a given graph. 

\begin{verbatim}
> ct + coord_cartesian(x = c(0.117, 0.120), y = c(-0.025, 0.025))
\end{verbatim}
 
\begin{figure}[ht]
	\centering
	\includegraphics[width=0.6\textwidth]{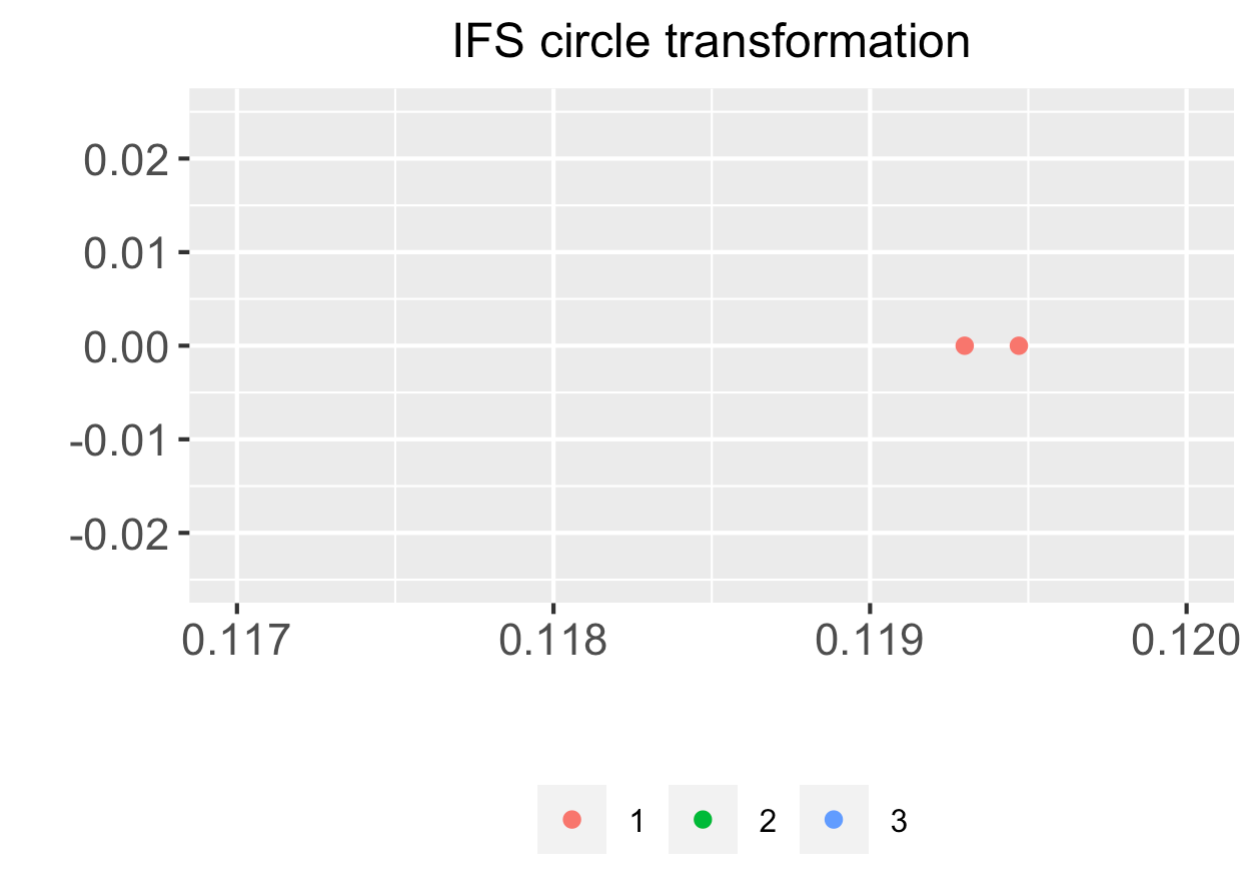}
	\caption{Region of interest (for subsequences of the form `111') concerning the circle transformation in the bottom right panel of Figure \ref{v1}.}
	\label{zoom}
\end{figure}

Figure~\ref{zoom} displays the resulting graph, which contains only two points. Therefore, we can conclude that only two subsequences of the form `111' exist in the first series in dataset \textit{SyntheticData1}. The frequency of any subsequence of arbitrary length could be obtained analogously. An interesting analysis regarding the interpretation of the graph resulting from the circle transformation can be seen in \cite{weiss2008visual} (Example 4.3). 

The previous graphs are frequently used for standard exploratory data analysis. However, \textbf{ctsfeatures} also contains some tools to carry out more specific tasks. One of such tools are the serial dependence plots described in Section~\ref{subsubsectionsdp}, which can be obtained by means of the functions \textit{plot\_cramers\_vi()} and \textit{plot\_cohens\_kappa()}. For instance, the plots showed in the left and right panels of Figure~\ref{v2} correspond to the values $\widehat{v}(l)$ and $\widehat{\kappa}(l)$, respectively, for the first series in  \textit{SyntheticData1}, and were obtained by running the code lines below.
\begin{verbatim}
> plot_1 <- plot_cramers_vi(series = SyntheticData1$data[[1]], 
+                           categories = categories)
> plot_2 <- plot_cohens_kappa(series = SyntheticData1$data[[1]], 
+                             categories = categories)
\end{verbatim}

By default, both functions consider lags from 1 to 10 (argument \textit{max\_lag}) and a significance level $\alpha=0.05$ for the tests in \eqref{test1} and \eqref{test2} (argument \textit{alpha}). As the standard autocorrelation plots, the corresponding estimates are displayed in a sequential order, with dashed lines indicating the critical values for the associated tests. In this case, the serial dependence plots clearly exhibit significant dependence at lags 1, 2 and 3. Moreover, dependence at lag 7 could also be considered significant, but this may be due to chance since multiple tests are simultaneously carried out. Indeed, the dependence structure suggested by both graphs is expected, since the considered series was generated from a MC. Certainly, a realization from a MC is expected to show a decreasing trend with respect to the degree of serial dependence. Specifically, the level of dependence is maximum at lag 1, then progressively decreases for further lags until it is no longer significant. Besides the dependence plots, functions \textit{plot\_cramers\_vi()} and \textit{plot\_cohens\_kappa()} also produce numerical outputs. For instance, the corresponding $p$-values can be obtained by using the argument \textit{plot = FALSE}. 
\begin{verbatim}
> plot_1 <- plot_cramers_vi(series = SyntheticData1$data[[1]], 
+                           categories = categories, plot = FALSE)
> plot_2 <- plot_cohens_kappa(series = SyntheticData1$data[[1]], 
+                             categories = categories, plot = FALSE)
> round(plot_1$p_values, 2) 
 [1] 0.00 0.00 0.00 0.28 0.88 0.61 0.01 0.05 0.07 0.07
> round(plot_2$p_values, 2) 
 [1] 0.00 0.00 0.00 0.07 0.62 0.15 0.01 0.24 0.04 0.04
\end{verbatim}
 
The $p$-values in these outcomes corroborate that both the Cramer\textquotesingle s $v$ and the Cohen\textquotesingle s $\kappa$ are significantly non-null at the first three lags, thus confirming the existence of significant serial dependence at these lags. Note that the $p$-value associated with lag 7 is rather low in both cases (0.01) and $\kappa(9)$ and $\kappa(10)$ also differ from zero in a significant way for $\alpha=0.05$. However, each set of $p$-values should be properly adjusted to handle random rejections of the null hypothesis that can arise in a multiple testing context. For instance, the well-known Holm\textquotesingle s method, which  controls the family-wise error rate at a pre-specified $\alpha$-level, could be applied to the $p$-values associated with the $\widehat{\kappa}(l)$-based test by executing
\begin{verbatim}
> p.adjust(round(plot_2$p_values, 2), method = `holm')
[1] 0.00 0.00 0.00 0.28 0.62 0.45 0.07 0.48 0.24 0.24
\end{verbatim}

According to the corrected $p$-values, the values of $\hat{\kappa}(l)$ for $l=1,2$ and 3 are still significant, but the null hypothesis of serial independence at lags 7, 9 and 10 cannot be now rejected.

\subsection{Performing data mining tasks with ctsfeatures}\label{subsectiondataminingtasks}

Jointly used with with external functions, \textbf{ctsfeatures} becomes a versatile and helpful tool to carry out different machine learning tasks involving categorical series. In this section, for illustrative purposes, we focus our attention on three important problems, namely classification, clustering and outlier detection.   


\subsubsection{Performing CTS classification}

Firstly, we show how the output of the functions in Table~\ref{tablefunctions2} can be used to perform feature-based classification. We illustrate this approach by considering the data collection \textit{GeneticSequences}, which contains the genome of 32 viruses pertaining to 4 different families (class labels). Using the functions \textit{gini\_index()} and \textit{cramers\_vi()}, we create the matrix \textit{feature\_dataset} formed by the values $\widehat g$ and $\widehat{v}(1)$ extracted from each one of the 32 series as follows. 
\begin{verbatim}
> categories2 <- factor(c(`a', `c', `g', `t'))
> list_1 <- lapply(GeneticSequences$data, gini_index, 
+                  categories = categories2)
> list_2 <- lapply(GeneticSequences$data, cramers_vi, lag=1, 
+                  categories = categories2)
> feature_dataset <- cbind(unlist(list_1), unlist(list_2))
> head(feature_dataset, 3)
          [,1]       [,2]
[1,] 0.9975936 0.11578926
[2,] 0.9971419 0.11874799
[3,] 0.9963633 0.11308114
\end{verbatim}

The $i$th row of \textit{feature\_dataset} contains estimated values characterizing the marginal and serial behavior of the $i$th CTS in the dataset. Therefore, several traditional classification algorithms can be applied to the matrix \textit{feature\_dataset} by means of the R package \textbf{caret} \cite{caret}. Package \textbf{caret} requires the dataset of features to be an object of class \textit{data.frame} whose last column must provide the class labels of the elements and be named \textit{`Class'}. Thus, as a preliminary step, we create \textit{df\_feature\_dataset}, a version of \textit{feature\_dataset} properly arranged to be used as input to \textbf{caret} functions, by means of the following chunk of code.
\begin{verbatim}
> df_feature_dataset <- data.frame(cbind(feature_dataset, 
+                                        GeneticSequences$classes))
> colnames(df_feature_dataset)[3] <- `Class'
> df_feature_dataset[,3] <- factor(df_feature_dataset[,3])
\end{verbatim}

The function \textit{train()} allows to fit several classifiers to the corresponding dataset, while the selected algorithm can be evaluated, for instance, by Leave-One-Out Cross-Validation (LOOCV). A grid search in the hyperparameter space of the corresponding classifier is performed by default. First we consider a standard classifier based on $k$ Nearest Neighbours ($k$NN) by using \textit{method = `knn'} as input parameter. By means of the command \textit{trControl()}, we define LOOCV as evaluation protocol.
\begin{verbatim}
> library(caret)
> train_control <- trainControl(method = `LOOCV')
> model_kNN <- train(Class~., data = df_feature_dataset, 
+                   trControl = train_control, method = `knn')
\end{verbatim}

The object \textit{model\_kNN} contains the fitted model and the evaluation results, among others. The reached accuracy can be accessed as follows.
\begin{verbatim}
> max(model_kNN$results$Accuracy)
[1] 0.78125
\end{verbatim}

The $k$NN classifier achieves an accuracy of 0.78 in the dataset of genetic sequences. Next we study the performance of the random forest. To this aim, we need to set \textit{method = `rf'}.
\begin{verbatim}
> model_RF <- train(Class~., data = df_feature_dataset, 
+                  trControl = train_control, method = `rf')
> max(model_RF$results$Accuracy)
[1] 0.90625
\end{verbatim}

The random forest obtains a much higher accuracy, around 0.91. Any other classifier can be evaluated in a similar way. Note that, in the previous analyses, the serial dependence structure of the series was summarized by means of a single estimate, $\widehat{v}(1)$. However, a more meaningful description of the CTS could be provided by employing the individual features involved in the definition of $\widehat{v}(l)$, that is the quantities $\frac{\big(\widehat{p}_{ij}(l)-\widehat{p}_i\widehat{p}_j\big)^2}{\widehat{p}_i\widehat{p}_j}$, $i,j=1,\ldots,r$. This way, the degree of deviation from serial independence is considered separately for each pair $(i,j)$. These features can be directly obtained by adding the argument \textit{features = TRUE} to the function \textit{cramers\_vi()}. To perform the classification by considering this new features, we should proceed as follows.
\begin{verbatim}
> list_2 <- lapply(GeneticSequences$data, cramers_vi, lag=1, 
+                  categories = categories2, features = T)
> new_feature_dataset <- cbind(unlist(list_1), matrix(unlist(list_2), 
+                           ncol = 16, byrow = T))
> dim(new_feature_dataset)
[1] 32 17
> df_new_feature_dataset <- data.frame(cbind(new_feature_dataset,
+                                            GeneticSequences$classes))
> colnames(df_new_feature_dataset)[18] <- `Class'
> df_new_feature_dataset[,18] <- factor(df_new_feature_dataset[,18])
> model_kNN <- train(Class~., data = df_new_feature_dataset, 
+                    trControl = train_control, method = `knn')
> max(model_kNN$results$Accuracy)
[1] 0.84375
> model_RF <- train(Class~., data = df_new_feature_dataset,
+                   trControl = train_control, method = `rf')
> max(model_RF$results$Accuracy)
[1] 0.90625
\end{verbatim}

While the $k$NN classifier improves its accuracy when the new features are considered, random forest exhibits exactly the same effectiveness for both sets of features.

Classification of CTS is often a challenging problem due to the complex nature of temporal data. If a feature-based approach is considered, the quality of the classification procedure strongly depends on the capability of the selected features to distinguish the underlying classes (as it has just been illustrated). In this context, trial and error (e.g., via LOOCV) is a common way of determining a suitable set of features. To illustrate the previous remarks, we decided to analyze the quality of the features based on pattern histograms (see Sections~\ref{subsubsectionph} and \ref{subsectionvisualizing}). Specifically, for each time series, we extract the counts of the cycle lengths for the category `a' (adenine). In particular, only the counts associated with cycles of lengths less than or equal to ten are considered. The following chunk of code creates the desired dataset. 
\begin{verbatim}`
> list_counts <- lapply(GeneticSequences$data, 
+        function(x) {pattern_histogram(x, 
+            category = `a', plot = F)[,2][1:10]})
> feature_counts <- matrix(unlist(list_counts), nrow = 32, byrow = T)
> head(feature_counts, 3)
     [,1] [,2] [,3] [,4] [,5] [,6] [,7] [,8] [,9] [,10]
[1,]  111   81   68   43   30   38   26   13   10     8
[2,]   99   77   65   48   31   31   25   13   17    11
[3,]  110   79   76   38   35   27   23   12   11     7
\end{verbatim}

Note that the argument \textit{plot = FALSE} in the function \textit{pattern\_histogram()} allows to obtain the counts. Let\textquotesingle s analyze the accuracy of the previously considered classifiers ($k$NN and random forest) with these new descriptive features. 
\begin{verbatim}
> df_feature_counts <- data.frame(cbind(feature_counts,
+                                GeneticSequences$classes))   
> colnames(df_feature_counts)[11] <- `Class'
> df_feature_counts[,11] <- factor(df_feature_counts[,11])
>  model_kNN  <- train(Class~., data = df_feature_counts,
+                   trControl = train_control, method = `knn')
> max(model_kNN$results$Accuracy)
[1] 0.78125
>  model_RF <- train(Class~., data = df_feature_counts,
+                 trControl = train_control, method = `rf')
> max(model_RF$results$Accuracy)
[1] 0.78125
\end{verbatim}

Although both methods show a worse performance than in previous analyses, they exhibit a rather high accuracy, thus indicating a great ability of the considered features (frequencies of cycle lengths for the category `a') to identify the underlying families of viruses. Indeed, we could examine the classification effectiveness of alternative sets of features (e.g., those involved in the definition of the estimates in Table \ref{tablefeaturescategoricalprocess}) by proceeding in exactly the same way.

\subsubsection{Performing CTS clustering}

The package \textbf{ctsfeatures} also provides an excellent framework to carry out clustering of categorical sequences. Let\textquotesingle s  consider the dataset \textit{SyntheticData1} and assume that the clustering structure is governed by the similarity between underlying models. In other terms, the ground truth is given by the 4 groups involving the 20 series from the same generating process. We wish to perform clustering and, according to our criterion, the clustering effectiveness of each algorithm must be measured by comparing the experimental solution with the true partition defined by these four groups.

In cluster analysis, distances between data objects play an essential role. In our case, a suitable metric should take low values for pairs of series coming from the same stochastic process, and high values otherwise. A classical exploratory step to shed light on the quality of a particular metric consists of constructing a two-dimensional scaling (2DS) based on the corresponding pairwise distance matrix. In short, 2DS represents the pairwise distances in terms of Euclidean distances into a 2-dimensional space preserving the original values as well as possible (by minimizing a loss function). For instance, we are going to construct the 2DS for dataset \textit{SyntheticData1} by using two specific metrics between CTS proposed by \cite{lopez2023hard} and denoted by $\widehat{d}_{CC}$ and $\widehat{d}_B$. More specifically, given two CTS, $\overline{X}_t^{(1)}$ and $\overline{X}_t^{(2)}$, the distances are defined as follows. 
\begin{equation}
\begin{split}
d_{CC}\big(\overline{X}_t^{(1)}, \overline{X}_t^{(2)}\big)= \sum_{l=1}^{L}\sum_{i,j=1}^{r}\Bigg(\frac{\big(\widehat{p}_{ij}^{(1)}(l)-\widehat{p}_i^{(1)}\widehat{p}_j^{(1)}\big)^2}{\widehat{p}_i^{(1)}\widehat{p}_j^{(1)}}-\frac{\big(\widehat{p}_{ij}^{(2)}(l)-\widehat{p}_i^{(2)}\widehat{p}_j^{(2)}\big)^2}{\widehat{p}_i^{(2)}\widehat{p}_j^{(2)}}\Bigg)^2\\
+\sum_{l=1}^{L}\sum_{i=1}^{r}\Bigg(\frac{\widehat{p}_{ii}^{(1)}(l)-\widehat{p}_i^{(1)}\widehat{p}_i^{(1)}}{1-\sum_{j=1}^{r}\widehat{p}_j^{(1)}\widehat{p}_j^{(1)}}-\frac{\widehat{p}_{ii}^{(2)}(l)-\widehat{p}_i^{(2)}\widehat{p}_i^{(2)}}{1-\sum_{j=1}^{r}\widehat{p}_j^{(2)}\widehat{p}_j^{(2)}}\Bigg)^2+\sum_{i=1}^{r}\big(\widehat{p}_i^{(1)}-\widehat{p}_i^{(2)}\big)^2,
\end{split}
\end{equation}
\begin{equation} \label{db}
d_B\big(\overline{X}_t^{(1)}, \overline{X}_t^{(2)}\big)=\sum_{l=1}^{L}\sum_{i,j=1}^{r}\big(\widehat{\psi}^{(1)}_{ij}(l)-\widehat{\psi}^{(2)}_{ij}(l)\big)^2+\sum_{i=1}^{r}\big(\widehat{p}_i^{(1)}-\widehat{p}_i^{(2)}\big)^2,
\end{equation} 
where $L$ denotes the largest lag and superscripts (1) and (2) indicate that the corresponding estimates are based on the realizations $\overline{X}_t^{(1)}$ and $\overline{X}_t^{(2)}$, respectively. Note that both dissimilarities assess discrepancies between the marginal distributions (last term) and the serial dependence structures (remaining terms) of both series. Therefore, they seem appropriate to group the CTS of a given collection in terms of underlying stochastic processes.

Let\textquotesingle s first create the datasets \textit{dataset\_dcc} and \textit{dataset\_db} with the features required to compute $d_{CC}$ and $d_{B}$, respectively. As the series in \textit{SyntheticData1} were generated from MC processes, we consider only one lag to construct the distance, i.e., we set $L=1$ (default option). We have to use the argument \textit{features = TRUE} in the corresponding functions.
\begin{verbatim}
> list_1_dcc <- lapply(SyntheticData1$data, cramers_vi,
+                      categories = factor(1:3), features = T)
> list_2_dcc <- lapply(SyntheticData1$data, cohens_kappa, 
+                      categories = factor(1:3), features = T)
> list_3_dcc <- lapply(SyntheticData1$data, marginal_probabilities,
+                      categories = factor(1:3))
> dataset_1_dcc <- matrix(unlist(list_1_dcc), nrow = 80, byrow = T)
> dataset_2_dcc <- matrix(unlist(list_2_dcc), nrow = 80, byrow = T)
> dataset_3_dcc <- matrix(unlist(list_3_dcc), nrow = 80, byrow = T)
> dataset_dcc <- cbind(dataset_1_dcc, dataset_2_dcc, dataset_3_dcc)
> list_1_db <- lapply(SyntheticData1$data, total_correlation,
+                     categories = factor(1:3), features = T)
> list_2_db <- lapply(SyntheticData1$data, marginal_probabilities,
+                     categories = factor(1:3))
> dataset_1_db <- matrix(unlist(list_1_db), nrow = 80, byrow = T)
> dataset_2_db <- matrix(unlist(list_2_db), nrow = 80, byrow = T)
> dataset_db <- cbind(dataset_1_db, dataset_2_db)
\end{verbatim}

The 2DS planes can be built using the function \textit{plot\_2d\_scaling()} of the R package \textbf{mlmts} \cite{mlmts}, which takes as input a pairwise dissimilarity matrix. 
\begin{verbatim}
> library(mlmts)
> distance_matrix_dcc <- dist(dataset_dcc)
> plot_dcc <- plot_2d_scaling(distance_matrix_dcc, 
+       cluster_labels = ctsfeatures::SyntheticData1$classes)$plot
> distance_matrix_db <- dist(dataset_db)
> plot_db <- plot_2d_scaling(distance_matrix_db,
+       cluster_labels = ctsfeatures::SyntheticData1$classes)$plot
\end{verbatim}

\begin{figure}[ht]
	\centering
	\includegraphics[width=0.7\textwidth]{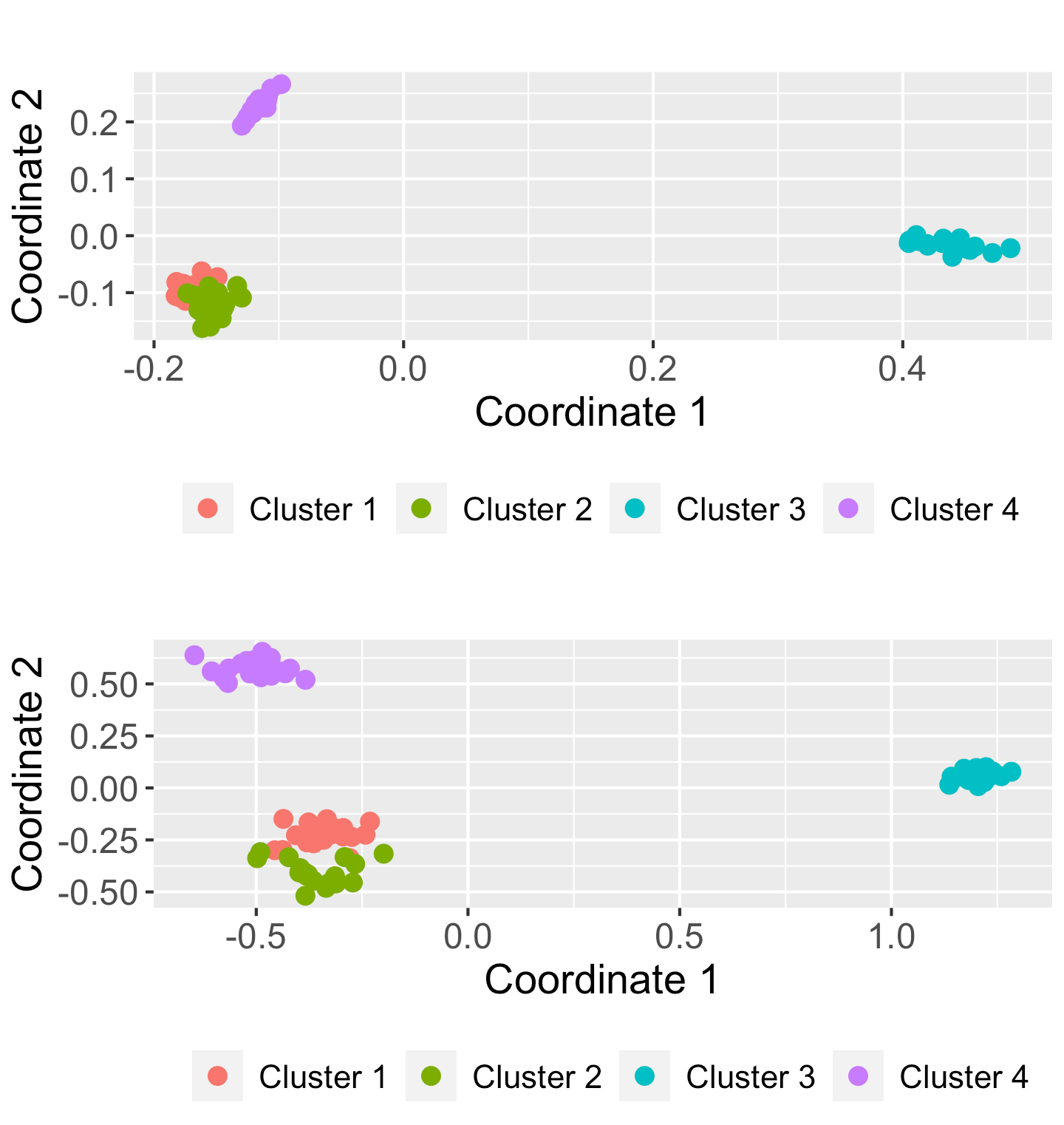}
	\caption{Two-dimensional scaling planes based on distances $d_{CC}$ (top panel) and $d_B$ (bottom panel) for the 80 series in the dataset \textit{SyntheticData1}.}
	\label{2dplotssd1}
\end{figure}

In the previous chunk of code, the syntax \textit{ctsfeatures::} was employed because package \textbf{mlmts} includes a data collection which is also called \textit{SyntheticData1}. The resulting plots are shown in Figure~\ref{2dplotssd1}. In both cases, the points were colored according to the true partition defined by the generating models. For it, we had to include the argument \textit{cluster\_labels} in the function \textit{plot\_2d\_scaling()}. This option is indeed useful to examine whether a specific metric is appropriate when the true class labels are known. The 2DS planes reveal that both metrics are able to identify the underlying structure rather accurately. However, there are two specific groups of CTS (the ones represented by red and green points) exhibiting a certain degree of overlap in both plots, which suggests a high level of similarity between the corresponding generating processes. This can be corroborated by observing that the coefficients in the transition matrices of these processes are certainly very similar (see Scenario~1 in Section~3.1 of \cite{lopez2023hard}). 

To evaluate the clustering accuracy of both metrics, we consider the popular Partitioning Around Medoids (PAM) algorithm, which is implemented in R through the function \textit{pam()} of package \textbf{cluster} \cite{cluster}. This function needs the pairwise distance matrix and the number of clusters. The latter argument is set to 4, since the series in dataset \textit{SyntheticData1} were generated from 4 different MC.
\begin{verbatim}
> library(cluster)
> clustering_dcc_pam <- pam(distance_matrix_dcc, k = 4)$clustering
> clustering_db_pam <- pam(distance_matrix_db, k = 4)$clustering
\end{verbatim}

The vectors \textit{clustering\_dcc} and \textit{clustering\_db} provide the respective clustering solutions based on both metrics. The quality of both partitions requires to measure their degree of agreement with the ground truth, which can be done by using the Adjusted Rand Index (ARI) \cite{campello2007fuzzy}. This index can be easily computed by means of the function \textit{external\_validation()} of package \textbf{ClusterR} \cite{clusterr}.
\begin{verbatim}
> library(ClusterR)
> external_validation(clustering_dcc_pam, 
+                       ctsfeatures::SyntheticData1$classes)
[1] 0.9662323
> external_validation(clustering_db_pam, 
+                       ctsfeatures::SyntheticData1$classes)
[1] 0.9038919
\end{verbatim}

The ARI index is bounded between -1 and 1 and admits a simple interpretation: the closer it is to 1, the better is the agreement between the ground truth and the experimental solution. Moreover, the value of 0 is associated with a clustering par- tition picked at random according to some simple hypotheses. Therefore, it can be concluded that both metrics $d_{CC}$ and $d_B$ attain great results in this dataset when used with the PAM algorithm. In particular, the partition produced by the former is slightly more similar to the ground truth. Note that the high value of ARI index was already expected from the 2DS plots in Figure \ref{2dplotssd1}. 

The popular $K$-means clustering algorithm can be also executed by using \textbf{ctsfeatures} utilities. In this case, we need to employ a dataset of features along with the \textit{kmeans()} function of package stats \cite{rsoftware}. 
\begin{verbatim}
> set.seed(123)
> clustering_dcc_kmeans <- kmeans(dataset_dcc, c = 4)$cluster
> external_validation(clustering_dcc_kmeans, 
+                     ctsfeatures::SyntheticData1$classes)
[1] 1
> set.seed(123)
> clustering_db_kmeans <- kmeans(dataset_db, c = 4)$cluster
> external_validation(clustering_db_kmeans, 
+                     ctsfeatures::SyntheticData1$classes)
[1] 0.9341667
\end{verbatim}

In this example, slightly better results are obtained when the $K$-means algorithm is employed, although the differences with respect to the ARI values produced by the PAM algorithm do not seem statistically significant. The performance of alternative dissimilarities or collections of features concerning a proper identification of the underlying clustering structure could be determined by following the same steps than in the previous experiments. 

\subsubsection{Performing outlier detection in CTS datasets}

Other challenging issue by analyzing a collection of CTS is to detect outlier elements. First, it is worthy noting that different notions of outlier are considered in the context of temporal data (additive outliers, innovative outliers, and others). Here, we consider the outlying elements to be
whole CTS objects. More specifically, an anomalous CTS is assumed to be a series generated from a stochastic process different from those generating the majority of the series in the database. 

To illustrate how \textbf{ctsfeatures} can be useful to carry out outlier identification, we create a dataset including two atypical elements. For it, we consider all the series in \textit{SyntheticData1} along with the first two series in dataset \textit{SyntheticData3}.  
\begin{verbatim}
> data_outliers <- c(SyntheticData1$data,SyntheticData3$data[1:2])
\end{verbatim}

The resulting data collection, \textit{data\_outliers}, contains 82 CTS. The first 80 CTS can be split into four homogeneous groups of 20 series, but those located into positions 81 and 82 are actually anomalous elements in the collection because they come from a NDARMA model (see Section~\ref{subsectiondatasets}). 

A distance-based approach to perform anomaly detection consists of obtaining the pairwise distance matrix and proceed in two steps as follows.
\begin{description}
\item[Step 1.] For each element, compute the sum of its distances from the remaining objects in the dataset, which is expected to be large for anomalous elements. 
\item[Step 2.] Sort the quantities computed in Step~1 in decreasing order and reordering the indexes according to this order. The first indexes in this new vector correspond to the most outlying elements, while the last ones to the least outlying elements.
\end{description}

We follow this approach to examine whether the outlying CTS in \textit{data\_outliers} can be identified by using the distance $d_B$ given in \eqref{db}. First, we construct the pairwise dissimilarity matrix based on this metric for the new dataset. 
\begin{verbatim}
> list_1_db_outl <- lapply(data_outliers, total_correlation,
+                              categories = factor(1:3), features = T)
> list_2_db_outl <- lapply(data_outliers, marginal_probabilities,
+                              categories = factor(1:3))            
> data_1_db_outl <- matrix(unlist(list_1_db_outl), nrow = 82, byrow = T)
> data_2_db_outl <- matrix(unlist(list_2_db_outl), nrow = 82, byrow = T)
> data_db_outl <- cbind(data_1_db_outl, data_2_db_outl)
> dist_db_outl <- dist(data_db_outl)
\end{verbatim}

Then, we apply the mentioned two-step procedure to matrix \textit{dist\_db\_outl} by running
\begin{verbatim}
> order(colSums(as.matrix(dist_db_outl)), decreasing = T)[1:2]
[1] 81 82
\end{verbatim}

Thus corroborating that $d_B$ is able to properly identify the two series generated from an anomalous stochastic process. As an illustrative exercise, let\textquotesingle s represent the corresponding 2DS plot for the dataset containing the two outlying CTS by using a different color for these elements. 
\begin{verbatim}
> library(mlmts)
> labels <- c(ctsfeatures::SyntheticData1$classes, 5, 5)
> plot_2d_scaling(dist_db_outl, cluster_labels = labels)$plot
\end{verbatim}

\begin{figure}[ht]
 	\centering
 	\includegraphics[width=0.7\textwidth]{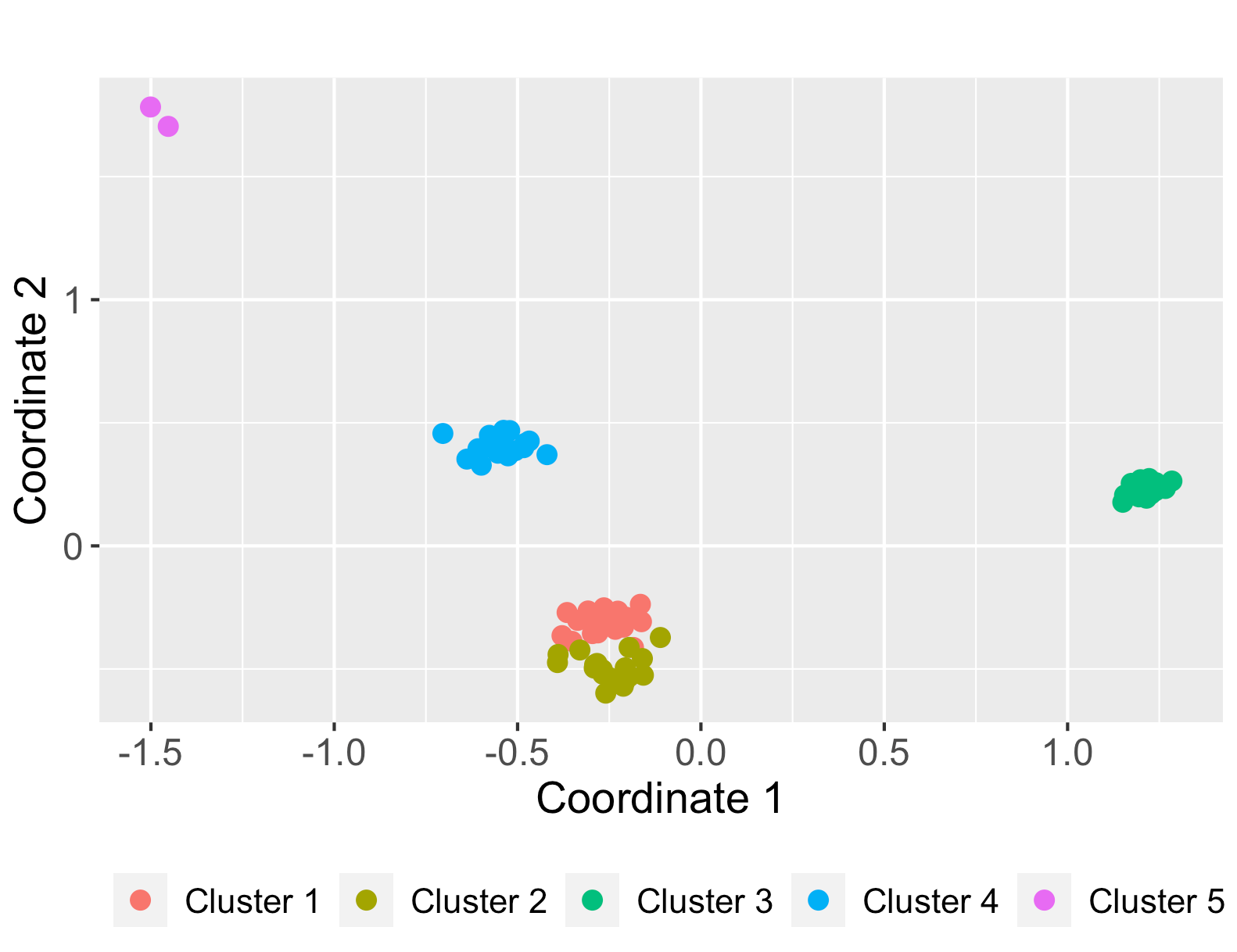}
 	\caption{Two-dimensional scaling plane based on distance $d_B$ (bottom panel) for the dataset containing 2 anomalous series.}
 	\label{2dplotoutliers}
 \end{figure}

The outcome is shown in Figure~\ref{2dplotoutliers}. The 2DS configuration is very similar to the one in the bottom panel of Figure~\ref{2dplotssd1} but, this time, two isolated points representing the anomalous series appear on the top left part of the plot. Clearly, 2DS plots can be very useful for outlier identification purposes, since they provide a great deal of information on both the number of potential outliers and their location with respect to the remaining elements in the dataset. 

In the previous example the number of outliers was assumed to be known, which is not realistic in practice. In fact, when dealing with real CTS datasets, one usually needs to determine whether the dataset at hand contains outliers. To that aim, it is often useful to define a measure indicating the outlying nature of each object (see, e.g., \cite{weng2008detecting} and \cite{lopez2021outlier}), i.e. those  elements with an extremely large scoring could be identified as outliers. In order to illustrate this approach, we consider the dataset \textit{GeneticSequences} and compute the pairwise distance matrix according to distance $d_B$. 
\begin{verbatim}
> list_3_db <- lapply(GeneticSequences$data, total_correlation,
+                 categories = factor(c(`a',`g',`c',`t')),features = T)
> list_4_db <- lapply(GeneticSequences$data, marginal_probabilities,
+                 categories = factor(c(`a',`g',`c',`t')))            
> data_3_db <- matrix(unlist(list_3_db), nrow = 32, byrow = T)
> data_4_db <- matrix(unlist(list_4_db), nrow = 32, byrow = T)
> data_db <- cbind(data_3_db, data_4_db)
> dist_db <- dist(data_db)
\end{verbatim} 

As before, the sum of the distances between each series and the remaining ones is computed. 
\begin{verbatim}
> outlier_score <- colSums(as.matrix(dist_db))
\end{verbatim}

The vector \textit{outlier\_score} contains the sum of the distances for each one of the 32 genetic sequences. Since the $i$th element of this vector can be seen as a measure of the outlying character of the $i$th sequence, those genetic sequences associated with extremely large values in this vector are potential outliers. A simple way to detect these series consists of visualizing a boxplot based on the elements of \textit{outlier\_score} and checking whether there are points located into the upper part of the graph. 
\begin{verbatim}
> boxplot(outlier_score, range = 1, col = `blue')
\end{verbatim}

\begin{figure}[ht]
	\centering
	\includegraphics[width=0.45\textwidth]{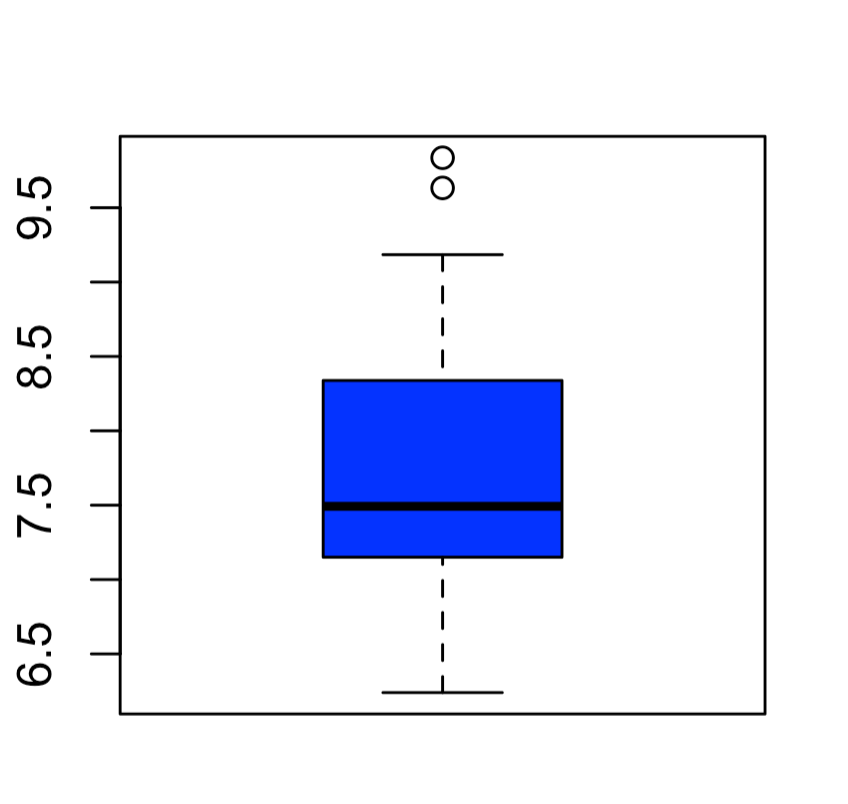}
	\caption{Boxplot of the outlying scores in dataset \textit{GeneticSequences} based on distance $d_B$. }
	\label{boxplot}
\end{figure}

The resulting boxplot is shown in Figure~\ref{boxplot} and suggests the existence of two series with abnormally high outlying scores. Hence, these genetic sequences could be considered to be anomalous and their individual properties be carefully investigated. Note that the prior empirical approach provides an automatic method to determine the number of outliers. Similar analyses could be carried out by considering alternative dissimilarity measures. 

\section{Concluding remarks}\label{sectionconcludingremarks}

Data mining of time series has experienced a significant growth during the 21st century. Although the majority of works focus on real-valued time series, categorical time series have received a great deal of attention during the last decade. The R package \textbf{ctsfeatures} is basically an attempt to provide different functions enabling to compute well-known statistical quantities for categorical series. Besides providing a valuable description about the structure of the time series, the corresponding features can be used as input for traditional machine learning procedures, as clustering, classification and outlier detection algorithms. Furthermore, \textbf{ctsfeatures} includes visualization tools and serial dependence plots, thus allowing the user to carry out valuable exploratory analyses. The main motivation behind the package is that, to the best of our knowledge, no previous R packages are available for a general statistical analysis of categorical series. In fact, the few software tools designed to deal with categorical sequences focus on specific tasks (e.g., computation of dissimilarities), application domains (e.g., sequence analysis), or types of categorical models (e.g., Markov Chains). Package \textbf{ctsfeatures} also incorporates two databases of biological sequences and several synthetic datasets, which can be used for illustrative purposes. Package \textbf{ctsfeatures} provides rather simple tools, but also necessary for standard analyses and of great usefulness to perform more complex tasks as modeling, inference, or forecasting. 

A description of the functions available in \textbf{ctsfeatures} is given in the first part of this work to make clear the details behind the software and its scope. However, the readers particularly interested in specific tools are encouraged to check the corresponding key references, which are also provided in the article. In the second part, the use of the package is illustrated by means of several examples involving synthetic and real data. This can be seen as a simple overview whose goal is to make the use of \textbf{ctsfeatures} as easy and understable as possible for first-time users. 

There are two main ways through which this work can be extended. First, as \textbf{ctsfeatures} is under continuous development, we expect to perform frequent updates by incorporating functions for the computation of additional statistical features which are introduced in the future. Second, the package focuses on the analysis of categorical series with nominal range. When the range is ordinal, alternative statistical features can be defined by taking into account the underlying ordering (see \cite{weiss2019distance}). Moreover, in the ordinal setting, these features are more meaningful than their nominal counterparts, since the former ignore the fact that a distance measure exists between the different categories. In this regard, the development of a software tool allowing the computation of statistical features for ordinal time series could be constructed in the future.

\section*{Acknowledgments}

This research has been supported by the Ministerio de Economía y Competitividad
(MINECO) grants MTM2017-82724-R and PID2020-113578RB-100, the Xunta de Galicia (Grupos de Referencia Competitiva ED431C-2020-14), and the Centro de Investigación del Sistema Universitariode Galicia,``CITIC" grant ED431G 2019/01; all of them through the European Regional Development Fund (ERDF).

\bibliography{mybibfile}

\begin{thebibliography}{10}
\expandafter\ifx\csname url\endcsname\relax
  \def\url#1{\texttt{#1}}\fi
\expandafter\ifx\csname urlprefix\endcsname\relax\def\urlprefix{URL }\fi
\expandafter\ifx\csname href\endcsname\relax
  \def\href#1#2{#2} \def\path#1{#1}\fi

\bibitem{weiss2014binomial}
C.~H. Wei{\ss}, P.~K. Pollett, Binomial autoregressive processes with
  density-dependent thinning, Journal of Time Series Analysis 35~(2) (2014)
  115--132.

\bibitem{stoffer2000spectral}
D.~S. Stoffer, D.~E. Tyler, D.~A. Wendt, The spectral envelope and its
  applications, Statistical Science (2000) 224--253.

\bibitem{fokianos2003regression}
K.~Fokianos, B.~Kedem, Regression theory for categorical time series,
  Statistical science 18~(3) (2003) 357--376.

\bibitem{weiss2008measuring}
C.~H. Wei{\ss}, R.~G{\"o}b, Measuring serial dependence in categorical time
  series, AStA Advances in Statistical Analysis 92 (2008) 71--89.

\bibitem{moysiadis2014binary}
T.~Moysiadis, K.~Fokianos, On binary and categorical time series models with
  feedback, Journal of Multivariate Analysis 131 (2014) 209--228.

\bibitem{krogh1994hidden}
A.~Krogh, M.~Brown, I.~S. Mian, K.~Sj{\"o}lander, D.~Haussler, Hidden markov
  models in computational biology: Applications to protein modeling, Journal of
  molecular biology 235~(5) (1994) 1501--1531.

\bibitem{weiss2018introduction}
C.~H. Wei{\ss}, An introduction to discrete-valued time series, John Wiley \&
  Sons, 2018.

\bibitem{cadez2003model}
I.~Cadez, D.~Heckerman, C.~Meek, P.~Smyth, S.~White, Model-based clustering and
  visualization of navigation patterns on a web site, Data mining and knowledge
  discovery 7 (2003) 399--424.

\bibitem{garcia2015framework}
M.~Garc{\'\i}a-Magari{\~n}os, J.~A. Vilar, A framework for dissimilarity-based
  partitioning clustering of categorical time series, Data mining and knowledge
  discovery 29~(2) (2015) 466--502.

\bibitem{lopez2023hard}
{\'A}.~L{\'o}pez-Oriona, J.~A. Vilar, P.~D’Urso, Hard and soft clustering of
  categorical time series based on two novel distances with an application to
  biological sequences, Information Sciences 624 (2023) 467--492.

\bibitem{li2022interpretable}
Z.~Li, S.~A. Bruce, T.~Cai, Interpretable classification of categorical time
  series using the spectral envelope and optimal scalings, Journal of Machine
  Learning Research 23~(299) (2022) 1--31.

\bibitem{horak2022nlp}
M.~Horak, S.~Chandrasekaran, G.~Tobar, Nlp based anomaly detection for
  categorical time series, in: 2022 IEEE 23rd International Conference on
  Information Reuse and Integration for Data Science (IRI), IEEE, 2022, pp.
  27--34.

\bibitem{rsoftware}
{R Core Team}, \href{https://www.R-project.org/}{R: A Language and Environment
  for Statistical Computing}, R Foundation for Statistical Computing, Vienna,
  Austria (2021).
\newline\urlprefix\url{https://www.R-project.org/}

\bibitem{confreq}
J.-H. Heine, R.~W. Alexandrowicz, M.~Stemmler,
  \href{https://CRAN.R-project.org/package=confreq}{{confreq}: Configural
  Frequencies Analysis Using Log-Linear Modeling}, r package version 1.6.1-1
  (2022).
\newline\urlprefix\url{https://CRAN.R-project.org/package=confreq}

\bibitem{heine2021analysis}
J.-H. Heine, M.~Stemmler, Analysis of categorical data with the r package
  confreq, Psych 3~(3) (2021) 522--541.

\bibitem{10.5555/1593511}
G.~Van~Rossum, F.~L. Drake, Python 3 Reference Manual, CreateSpace, Scotts
  Valley, CA, 2009.

\bibitem{gabadinho2011analyzing}
A.~Gabadinho, G.~Ritschard, N.~S. Mueller, M.~Studer, Analyzing and visualizing
  state sequences in r with traminer, Journal of statistical software 40~(4)
  (2011) 1--37.

\bibitem{studer2016matters}
M.~Studer, G.~Ritschard, What matters in differences between life trajectories:
  A comparative review of sequence dissimilarity measures, Journal of the Royal
  Statistical Society. Series A (Statistics in Society) (2016) 481--511.

\bibitem{liao2022sequence}
T.~F. Liao, D.~Bolano, C.~Brzinsky-Fay, B.~Cornwell, A.~E. Fasang, S.~Helske,
  R.~Piccarreta, M.~Raab, G.~Ritschard, E.~Struffolino, et~al., Sequence
  analysis: Its past, present, and future, Social science research 107 (2022)
  102772.

\bibitem{statacorpstata}
L.~StataCorp, Stata statistical software: Release 17. 2021, College Station,
  Texas, United States of America.

\bibitem{halpin2017sadi}
B.~Halpin, Sadi: Sequence analysis tools for stata, The Stata Journal 17~(3)
  (2017) 546--572.

\bibitem{stoffer1993spectral}
D.~S. Stoffer, D.~E. Tyler, A.~J. McDougall, Spectral analysis for categorical
  time series: Scaling and the spectral envelope, Biometrika 80~(3) (1993)
  611--622.

\bibitem{shumway2000time}
R.~H. Shumway, D.~S. Stoffer, D.~S. Stoffer, Time series analysis and its
  applications, Vol.~3, Springer, 2000.

\bibitem{fitzgerald2004clustering}
P.~C. FitzGerald, A.~Shlyakhtenko, A.~A. Mir, C.~Vinson, Clustering of dna
  sequences in human promoters, Genome research 14~(8) (2004) 1562--1574.

\bibitem{kassim2017classification}
N.~A. Kassim, A.~Abdullah, Classification of dna sequences using convolutional
  neural network approach, UTM Computing Proceedings Innovations in Computing
  Technology and Applications 2 (2017) 1--6.

\bibitem{weiss2008visual}
C.~H. Wei{\ss}, Visual analysis of categorical time series, Statistical
  Methodology 5~(1) (2008) 56--71.

\bibitem{ribler1997visualizing}
R.~L. Ribler, Visualizing categorical time series data with applications to
  computer and communications network traces, Ph.D. thesis, Virginia
  Polytechnic Institute and State University (1997).

\bibitem{weiss2005discover}
C.~H. Wei{\ss}, R.~G{\"o}b, Discover Patterns in Categorical Time Series Using
  IFS, Inst. of Applied Math. and Statistics, 2005.

\bibitem{weiss2013serial}
C.~H. Wei{\ss}, Serial dependence of ndarma processes, Computational Statistics
  \& Data Analysis 68 (2013) 213--238.

\bibitem{weiss2011empirical}
C.~H. Wei{\ss}, Empirical measures of signed serial dependence in categorical
  time series, Journal of Statistical Computation and Simulation 81~(4) (2011)
  411--429.

\bibitem{astsa}
David, N.~Poison, \href{https://CRAN.R-project.org/package=astsa}{astsa:
  Applied Statistical Time Series Analysis}, r package version 1.16 (2022).
\newline\urlprefix\url{https://CRAN.R-project.org/package=astsa}

\bibitem{ggplot2}
H.~Wickham, \href{https://ggplot2.tidyverse.org}{ggplot2: Elegant Graphics for
  Data Analysis}, Springer-Verlag New York, 2016.
\newline\urlprefix\url{https://ggplot2.tidyverse.org}

\bibitem{caret}
M.~Kuhn, \href{https://CRAN.R-project.org/package=caret}{caret: Classification
  and Regression Training}, r package version 6.0-93 (2022).
\newline\urlprefix\url{https://CRAN.R-project.org/package=caret}

\bibitem{mlmts}
A.~Lopez-Oriona, J.~{A. Vilar},
  \href{https://CRAN.R-project.org/package=mlmts}{mlmts: Machine Learning
  Algorithms for Multivariate Time Series}, r package version 1.1.1 (2023).
\newline\urlprefix\url{https://CRAN.R-project.org/package=mlmts}

\bibitem{cluster}
M.~Maechler, P.~Rousseeuw, A.~Struyf, M.~Hubert, K.~Hornik,
  \href{https://CRAN.R-project.org/package=cluster}{cluster: Cluster Analysis
  Basics and Extensions}, r package version 2.1.2 --- For new features, see the
  'Changelog' file (in the package source) (2021).
\newline\urlprefix\url{https://CRAN.R-project.org/package=cluster}

\bibitem{campello2007fuzzy}
R.~J. Campello, A fuzzy extension of the rand index and other related indexes
  for clustering and classification assessment, Pattern Recognition Letters
  28~(7) (2007) 833--841.

\bibitem{clusterr}
L.~Mouselimis, \href{https://CRAN.R-project.org/package=ClusterR}{{ClusterR}:
  Gaussian Mixture Models, K-Means, Mini-Batch-Kmeans, K-Medoids and Affinity
  Propagation Clustering}, r package version 1.2.6 (2022).
\newline\urlprefix\url{https://CRAN.R-project.org/package=ClusterR}

\bibitem{weng2008detecting}
X.~Weng, J.~Shen, Detecting outlier samples in multivariate time series
  dataset, Knowledge-based systems 21~(8) (2008) 807--812.

\bibitem{lopez2021outlier}
{\'A}.~L{\'o}pez-Oriona, J.~A. Vilar, Outlier detection for multivariate time
  series: A functional data approach, Knowledge-Based Systems 233 (2021)
  107527.

\bibitem{weiss2019distance}
C.~H. Wei{\ss}, Distance-based analysis of ordinal data and ordinal time
  series, Journal of the American Statistical Association (2019).

\end{thebibliography}

\end{document}